\documentclass[sigconf,nonacm]{acmart}

\usepackage{amsmath}
\usepackage{algorithm,algpseudocode}
\usepackage{booktabs,graphicx,xcolor}
\usepackage{array,tabularx,ragged2e,adjustbox}
\usepackage{microtype}
\newcolumntype{Y}{>{\RaggedRight\arraybackslash}X}
\definecolor{revisionred}{RGB}{185,28,38}
\newcommand{\framework}{ReSemReason}
\newcommand{\coremethod}{ReSemAttn}

\newcommand{\biohopr}{BioHopR}

\setcopyright{none}
\acmDOI{}\acmISBN{}
\acmConference[WWW '27]{The ACM Web Conference 2027}{May 10--14, 2027}{Dublin, Ireland}
\acmYear{2027}\copyrightyear{2027}

\title{Beyond Reachability: A Relation-Semantic Reasoning
Framework for Biomedical Multi-hop QA}

\author{Ziyang Ling}
\email{zaneling@mail.ustc.edu.cn}
\affiliation{
  \institution{University of Science and Technology of China}
  \city{Hefei}
  \country{China}
}

\author{Ziyi Chen}
\email{chenziyi@mail.ustc.edu.cn}
\affiliation{
  \institution{University of Science and Technology of China}
  \city{Hefei}
  \country{China}
}

\author{Ronald X.Xu}
\email{chenziyi@mail.ustc.edu.cn}
\affiliation{
  \institution{University of Science and Technology of China}
  \city{Hefei}
  \country{China}
}

\author{Mingzhai Sun}
\email{chenziyi@mail.ustc.edu.cn}
\affiliation{
  \institution{University of Science and Technology of China}
  \city{Hefei}
  \country{China}
}

\begin{document}
\begin{abstract}
Biomedical multi-hop reasoning over knowledge graphs requires identifying a semantically valid relation sequence that supports the answer. Existing methods can retrieve correct answers, yet often rank them poorly within a large pool of seemingly relevant candidates. Even when a correct answer is retrieved, its associated evidence and reasoning path may fail to support the relation composition required by the question. We term this disconnect between answer reachability and evidential validity the \textbf{reachable-but-invalid} problem. We present \textbf{\framework{}}, which shares a question-induced distribution over typed, directed relation schemas between candidate discovery and evidence selection. Schema compatibility guides graph search; \textbf{\coremethod{}} then uses relation-conditioned key/value projections and an uncertainty-weighted path prior to attend to candidate-specific evidence. A separate absolute-support score complements normalized attention when all retrieved paths are weak. The reported experiments on BioHopR, PrimeKGQA, and MedReason show improvements in answer ranking and answer-set quality. Path-discrimination analyses examine the distinction between endpoint correctness and evidential support. We also specify matched controls for search, attention, and schema uncertainty. Code: \url{https://anonymous.4open.science/r/ReSemAttn}.
\end{abstract}
\maketitle

\section{Introduction}
\label{sec:introduction}

Biomedical multi-hop reasoning over knowledge graphs requires identifying a semantically valid relation sequence that supports the answer. A question may connect a drug to its target proteins and then to associated diseases, requiring several facts to be composed under specific relational constraints. The knowledge needed for such reasoning is distributed across scientific publications and biomedical databases available on the Web. Knowledge graphs such as PrimeKG integrate heterogeneous resources into typed entities and directed relations~\citep{chandak2023primekg}, making this knowledge accessible through a shared, machine-interpretable structure. For biomedical search and question answering, effective access to this structure requires interpreting how relations compose into evidence for a user's question.

\begin{figure}[!tbp]
\centering
\includegraphics[width=\columnwidth]{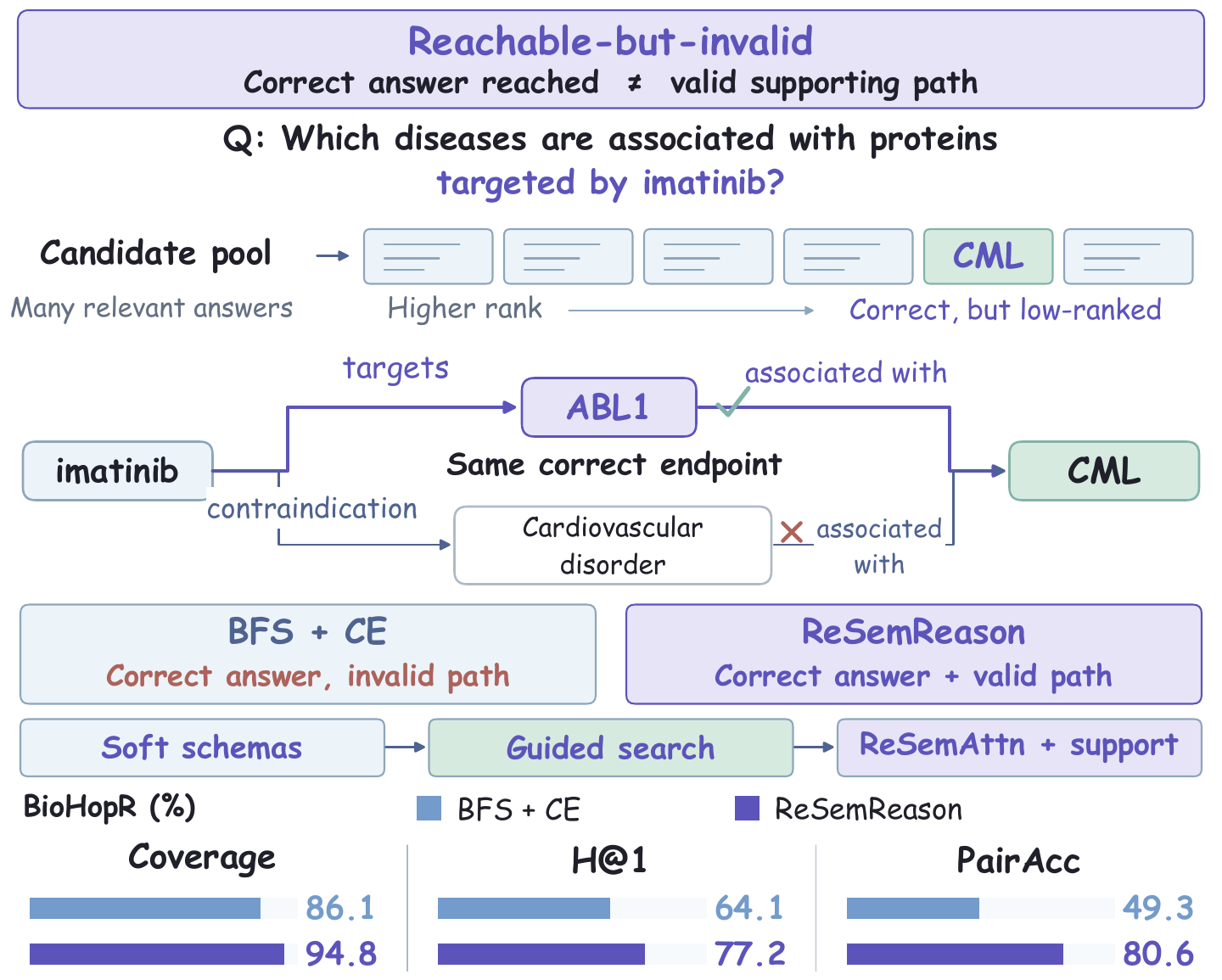}
\caption{Two paths reach correct answer chronic myeloid leukemia (CML), but only the ABL1 path supports the question. Bottom: reported BioHopR Coverage, H@1, and same-answer PairAcc.}
\Description{A single-column motivation figure contrasts a target-protein path through ABL1 with a contraindication path through cardiovascular disorder; both reach CML. The baseline favors invalid support, whereas ReSemReason favors the ABL1 path. Three connected blocks show soft schemas, guided search, and ReSemAttn with absolute support. Bar charts compare Coverage 86.1 versus 94.8 percent, H at 1 of 64.1 versus 77.2 percent, and same-answer PairAcc of 49.3 versus 80.6 percent.}
\label{fig:motivation}
\end{figure}

This creates a semantic knowledge-access challenge for the Web: how can systems retrieve and combine interlinked biomedical facts while respecting the relation composition requested by a user? A path can connect relevant entities, and even reach the correct answer, without providing the required evidence. We call this the \textbf{reachable-but-invalid problem}. As illustrated in Figure~\ref{fig:motivation}, distinguishing valid support from misleading connectivity requires attention to relation types, traversal directions, and intermediate entity types. We address this challenge by using question-induced relation schemas to guide both candidate discovery and evidence selection. Our contribution concerns the semantic consumption of integrated biomedical knowledge which making explicit relation semantics operational in retrieval and reasoning, and assessing supporting paths alongside answer correctness. This connects biomedical multi-hop reasoning to the broader goal of reliable knowledge access for Web search, question answering, and knowledge-grounded agents.

Recent research has increasingly combined large language models (LLMs) with structured biomedical knowledge. A survey of biomedical and healthcare LLM--KG integration identifies KG-guided reasoning and verification as important directions for improving factual grounding and interpretability~\citep{xu2025biomedkg}. More broadly, a recent survey of retrieval-augmented generation and reasoning emphasizes their complementary roles that retrieval supplies external premises~\citep{li2025ragreasoning}. This interaction is particularly relevant to biomedical questions, where several individually relevant facts may still fail to form an appropriate answer-supporting chain. Benchmarks and methods develop this interaction from different perspectives. BioHopR evaluates multi-hop, multi-answer biomedical reasoning~\citep{kim2025biohopr}, while MedReason uses knowledge graphs to construct grounded medical reasoning data~\citep{wu2025medreason}. More recently, eGoT combines iterative reasoning with biomedical knowledge retrieval to build evidence graphs for question answering and hypothesis generation~\citep{sanda2026egot}. In general KG question answering, RoG uses relation paths to organize planning, retrieval, and reasoning~\citep{luo2024rog}; GNN-RAG uses graph neural retrieval to identify relevant candidates and extract paths for LLM reasoning~\citep{mavromatis2025gnnrag}; and GCR constrains decoding with a KG-derived path index~\citep{luo2025gcr}. These advances establish the value of structured evidence and relation-aware reasoning.

These advances motivate examining both answer ranking and the evidence supporting a prediction. A correct answer may be retrieved within a large pool of seemingly relevant candidates yet remain poorly ranked. Even when ranking succeeds, answer accuracy alone cannot distinguish valid supporting evidence from a semantically incompatible path to the same endpoint. Evaluating the complete relation composition is therefore necessary to assess whether the retrieved evidence actually answers the question. Consider the running question, \emph{Which diseases are associated with proteins targeted by imatinib?} As shown in Figure~\ref{fig:motivation}, the path from imatinib through ABL1 to chronic myeloid leukemia (CML) follows the requested drug--target-protein--disease composition. A contraindication branch reaches the same CML endpoint without establishing the required target-protein connection. Holding the answer entity fixed isolates the evidence-selection challenge: the system must distinguish these paths by their relation semantics.

Addressing this problem requires coordinating two stages. During candidate discovery, semantically incompatible branches can consume a limited search budget and displace valid evidence. A downstream reranker cannot recover candidates or paths that retrieval has already discarded. During evidence selection, several paths to the same candidate may compete for influence, although only some match the requested relation composition. Moreover, normalized attention expresses relative preference among retrieved paths: it can concentrate on one path even when every available path provides weak support. Candidate availability, relative evidence preference, and absolute support therefore need to be distinguished.

Question interpretation introduces an additional difficulty. Biomedical terminology and relation descriptions can admit several plausible schemas. Committing to one predicted schema can prematurely suppress a valid branch, whereas ignoring schema information leaves search and reasoning without an explicit representation of the requested composition. This motivates retaining a distribution over typed, directed relation schemas and using it consistently across the two stages, while accounting for uncertainty when it influences evidence attention.

\begin{figure*}[!t]
\centering
\includegraphics[width=\textwidth]{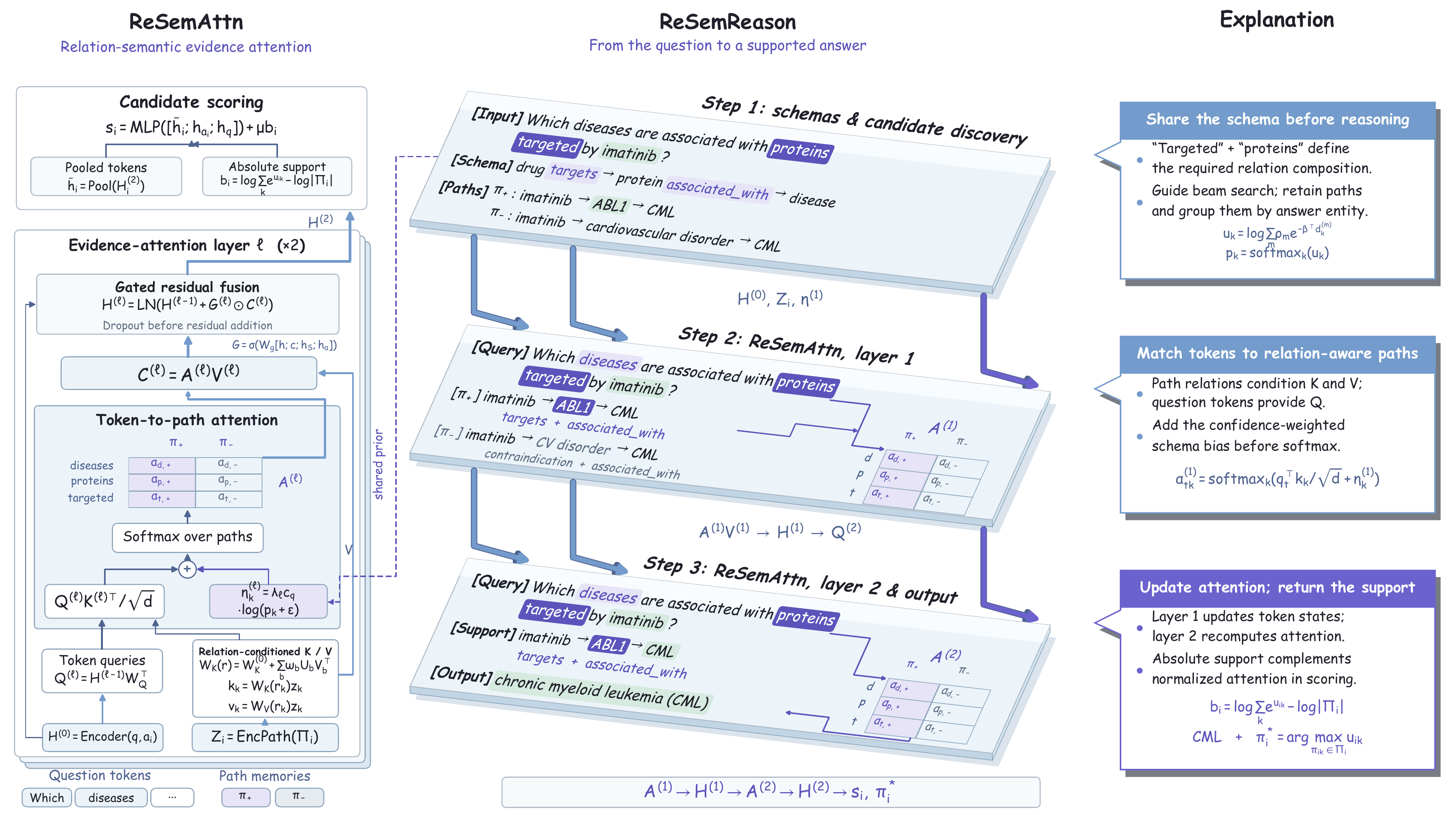}
\caption{Overview of \framework{} pipeline. Left: the two-layer \coremethod{} computation. Middle: input, schema-guided discovery, attention updates, and the CML answer with its ABL1 support. Right: shared schemas, relation-conditioned attention, and absolute support.}
\Description{A detailed ReSemAttn block on the left shows relation-conditioned key and value projections, token queries, a shared schema bias, softmax over paths, value aggregation, and gated residual fusion. Three inclined case planes in the center progress from the imatinib question and candidate paths to the first and second attention layers, ending with chronic myeloid leukemia and the ABL1 path. Right-side explanations connect each stage to the method's innovations. The matrices show symbolic token-path weights and do not encode measured attention magnitudes.}
\label{fig:framework}
\end{figure*}

We present \textbf{\framework{}}, a relation-semantic reasoning framework that shares a question-induced schema distribution between candidate discovery and neural evidence selection. As shown in Figure~\ref{fig:framework}, the framework first predicts several plausible schemas specifying relation sequences, traversal directions, and entity types. Their compatibility with partial paths guides graph search, prioritizing relevant branches while preserving alternative interpretations through soft penalties. The resulting candidate-specific paths form the evidence memory for \textbf{\coremethod{}}. Relation-conditioned key/value projections shape how paths are represented, and an uncertainty-weighted compatibility prior guides token-to-path attention. Finally, an absolute-support term complements normalized attention, retaining information about whether the retrieved paths provide strong support at all. Together, these components connect the semantics used to discover evidence with those used to incorporate it into answer prediction.

Our contributions are (1) We formulate the reachable-but-invalid problem, distinguishing candidate availability and answer correctness from whether a supporting path satisfies the question's relation composition. (2) We introduce ReSemReason, which shares soft relation schemas between candidate discovery and evidence selection, and ReSemAttn, which combines relation-conditioned representations, uncertainty-weighted attention, and absolute support.

\section{Related Work}
\label{sec:related_work}

\noindent\textbf{Biomedical multi-hop reasoning.}
MedHop studies evidence composition across biomedical documents~\citep{welbl2018constructing}. More recent resources emphasize structured reasoning: BioHopR evaluates multi-answer reasoning over PrimeKG paths, PrimeKGQA provides executable graph queries, and MedReason constructs KG-grounded medical reasoning data~\citep{kim2025biohopr,primekgqa2024zenodo,wu2025medreason}. Recent work further explores biomedical LLM--KG integration~\citep{xu2025biomedkg}, including iterative evidence retrieval and graph-based reasoning in eGoT~\citep{sanda2026egot}.

\noindent\textbf{Retrieval and graph reasoning.}
Earlier approaches combine question-conditioned retrieval with graph and language representations~\citep{sun2019pullnet,yasunaga2021qagnn,zhang2022greaselm,jiang2023unikgqa}. Recent methods advance iterative graph exploration, relation-path planning, graph neural retrieval, and graph-constrained decoding~\citep{sun2024think,luo2024rog,mavromatis2025gnnrag,luo2025gcr}, reflecting the growing integration of retrieval and reasoning~\citep{li2025ragreasoning}. ReSemReason shares soft, typed, directed relation schemas across candidate discovery and evidence selection.
\section{Method}\label{sec:method}
\subsection{Task and Framework}\label{subsec:problem_definition}\label{subsec:overview}
Given question $q$ and directed KG $\mathcal G=(\mathcal V,\mathcal R,\mathcal E)$, the system discovers candidate entities $\mathcal C_q$, ranks them, and returns an answer set $\widehat{\mathcal A}_q$ with supporting paths. For a path $\pi$, three events differ: the path exists in the KG; its endpoint belongs to the reference set $\mathcal A_q$; and its relation composition supports that endpoint as an answer to $q$. We write the third event as $\mathrm{Valid}(q,\pi)$ and evaluate it independently of the predicted schema.

For answer $a_i$, the selected explanation $\pi_i^*$ must end at $a_i$. Success thus requires both \emph{availability}, meaning that suitable candidates and evidence survive retrieval, and \emph{selection}, meaning that the reasoner uses them correctly. Figure~\ref{fig:framework} follows the same imatinib case through grounding and schema prediction, candidate discovery, and path-based reasoning. Appendix Sections~A--C provide the evaluation and implementation details.

\noindent\textbf{An availability ceiling.}
For methods whose output is restricted to $\mathcal C_q$, let $A_q$ denote top-answer correctness and $G_q=\mathbf1[\mathcal A_q\cap\mathcal C_q\ne\varnothing]$. Then $A_q\leq G_q$, giving
\begin{equation}
 \mathrm{H@1}=\mathrm{Coverage}\cdot\mathrm{Cond.H@1},
 \qquad \mathrm{H@1}\leq\mathrm{Coverage},
 \label{eq:availability_ceiling}
\end{equation}
when conditional accuracy is defined on questions with $G_q=1$. This identity is per run on a common question set. It explains why reranking cannot compensate for missing candidates and motivates reporting both terms in E1. For multi-answer questions, Coverage is weaker than candidate recall: finding one answer can satisfy the former while leaving most of the answer set unavailable.

\subsection{Grounding and Question-Only Schemas}\label{subsec:question_understanding}
We retain the top-$N$ anchor entities under $p_\phi(e_0\mid q)$, rather than committing to a single lexical match. The reference grounder applies a softmax to cosine similarities between pooled question and entity-text representations, with temperature $T_g=0.1$. It scores the loaded entity inventory before truncation. A schema describes hop count, entity types, relation families, and directions:
\begin{equation}
 S=(L,\tau_0,r_1,d_1,\tau_1,\ldots,r_L,d_L,\tau_L).
 \label{eq:schema_main_revised}
\end{equation}
Independent heads over the pooled question representation score these components. A bounded enumeration retains the top-$M$ hypotheses, whose scores $\ell_m$ are renormalized:
\begin{equation}
 \rho_m=\frac{\exp\ell_m}{\sum_{j=1}^{M}\exp\ell_j},\qquad
 \mathcal S_q=\{S_q^{(m)}\}_{m=1}^{M}.
 \label{eq:soft_schema_main_revised}
\end{equation}
The inference interface uses the question rather than gold test schemas. Schema enumeration and the model configuration are specified in Appendix Sections~B.2 and~C.

The confidence gate measures concentration within the retained hypotheses:
\begin{equation}
 c_q=\begin{cases}1,&M=1,\\
 1-H(\boldsymbol\rho)/\log M,&M>1,
 \end{cases}\quad
 H(\boldsymbol\rho)=-\sum_m\rho_m\log\rho_m.
 \label{eq:schema_confidence}
\end{equation}
Uniform hypotheses give $c_q=0$. This gate does not measure whether the correct schema survived truncation. For the running case, keeping several interpretations allows a target-protein branch to remain competitive even when it is not associated with the highest-scoring schema.

\noindent\textbf{Uncertainty has two sources.}
High entropy can reflect several plausible interpretations, whereas low entropy can still reflect a confidently wrong retained set. The gate cannot recover a schema omitted from the top-$M$ hypotheses. Appendix Section~\ref{app:schema_calibration} evaluates confidence calibration for hop, relation-sequence, and full-schema prediction.

\subsection{Relation-Guided Candidate Discovery}\label{subsec:relation_guided_search}
Starting from each retained anchor, search expands directed edges and normalized inverse traversals. For an extension $\pi'$ ending at $e'$, its score is
\begin{equation}
\begin{aligned}
 s_{\mathrm{search}}(\pi')={}&\lambda_g\log p_\phi(e_0\mid q)
 +\lambda_s\operatorname{Compat}(q,\pi')\\
 &+\lambda_q\operatorname{Sem}(q,e')-\lambda_hh(e'),\\
 \operatorname{Compat}(q,\pi')={}&
 \log\sum_m\rho_m\exp[-D_{\mathrm{prefix}}(S_q^{(m)},\pi')].
\end{aligned}\label{eq:search_score_main_revised}
\end{equation}
Here $D_{\mathrm{prefix}}$ sums position-aligned relation, direction, and destination-type mismatches over the overlapping prefix. Endpoint relevance is cosine similarity and $h(e')=\log(1+\deg(e'))$. The reference coefficients are $(1,2,0.5,0.15)$; the explicit mismatch and search procedure appear in Appendix Section~B.3.

Search rejects repeated nodes and retains the top-$B$ continuing paths at each depth. Paths reaching a retained schema's hop count enter the completed collection before the next beam is pruned. Completed paths are grouped by endpoint, with at most $K$ retained per candidate. Thus $\mathcal C_q$ is discovered during traversal rather than supplied to a reranker. Compatibility penalties are soft: matching a lower-probability schema can preserve a branch, although finite budgets and grounding errors can still remove valid evidence. Search is discrete and uses fixed coefficients in the reference implementation; \coremethod{} does not feed scores back into expansion.

\noindent\textbf{Following the running case.}
The target-protein branch matches the dominant schema; the contraindication branch mismatches both relation and bridge type. The protein-to-disease hop then completes the requested composition. Endpoint relevance alone cannot distinguish these routes. Soft mismatch penalties change search priority before pruning, while independent annotation determines whether the completed evidence is valid.

\subsection{\coremethod{}: Relation-Semantic Path Attention}\label{subsec:resemattn_main_revised}
For candidate $a_i$, let $\Pi_i=\{\pi_{ik}\}$ be its retained paths. Each memory combines entity-type, directed-relation, path-summary, and topology features:
\begin{equation}
\begin{aligned}
 f_{ik}&=[\phi_E(E_{ik});\phi_R(R_{ik});\phi_P(P_{ik});\phi_T(T_{ik})],\\
 z_{ik}&=\operatorname{LN}\bigl(\operatorname{Dropout}(\operatorname{GELU}(W_zf_{ik}))\bigr).
\end{aligned}\label{eq:path_memory_main_revised}
\end{equation}
The reference encoder pools type and relation embeddings and uses numerical path and degree features. Relation order is compared explicitly in the schema mismatch, rather than preserved by mean pooling. These features do not estimate edge-provenance reliability.

\noindent\textbf{Semantic compatibility.}
For each schema, the mismatch vector contains type disagreement, directed-relation disagreement, length/closure mismatch, and mean log-degree. With nonnegative weights $\boldsymbol\beta=\operatorname{softplus}(\boldsymbol\gamma)$,
\begin{equation}
\begin{aligned}
 \mathcal E_{ik}^{(m)}&=\boldsymbol\beta^\top\mathbf d_{ik}^{(m)},\\
 u_{ik}&=\log\sum_m\rho_m\exp(-\mathcal E_{ik}^{(m)}),\qquad
 p_{ik}=\operatorname{softmax}_k(u_{ik}).
\end{aligned}\label{eq:path_prior_main_revised}
\end{equation}
For a path of length $L$, let $v_j$ be its $j$th entity and $(r_j,d_j)$ its incoming relation and direction. Barred symbols denote a schema of length $H$. Define alignment lengths $J=\min(L,H)$ and $J_e=\min(L+1,H+1)$. The mismatch vector is
\begin{equation}
\begin{aligned}
 D_E&=J_e^{-1}\sum_{j=0}^{J_e-1}\mathbf1[\tau(v_j)\ne\bar\tau_j],\\
 D_R&=\frac{\sum_{j=1}^{J}\bigl(\mathbf1[r_j\ne\bar r_j]
             +0.75\mathbf1[d_j\ne\bar d_j]\bigr)}{\max(1,J)},\\
 D_P&=|L-H|+\mathbf1[v_0=v_L],\\
 D_T&=(L+1)^{-1}\sum_{j=0}^{L}\log(1+\deg(v_j)),\\
 \mathbf d&=[D_E,D_R,D_P,D_T]^\top.
\end{aligned}\label{eq:semantic_factors}
\end{equation}
Entity matching compares types; directed-relation matching preserves positional order. The path term penalizes length and closure, and the topology term measures graph degree rather than provenance quality. Since search rejects repeated nodes, closure is normally zero. Nonnegative energies and normalized schema weights imply $u_{ik}\leq0$; increasing a mismatch cannot improve compatibility. This makes the score inspectable without equating schema agreement with independently annotated validity.

Search sums mismatches over incomplete prefixes; the evidence prior normalizes aligned type/relation terms for completed paths. They share schema content and directionality, while scoring different decisions: which extension to explore and how much a retained explanation should contribute.

\noindent\textbf{Relation-conditioned representation and selection.}
Each path's relation embedding conditions low-rank mixtures for keys and values,
\begin{equation}
 k_{ik}^{\ell}=W_K^{\ell}(r_{ik})z_{ik},\qquad
 v_{ik}^{\ell}=W_V^{\ell}(r_{ik})z_{ik}.
 \label{eq:relation_kv_main_revised}
\end{equation}
For the pooled relation representation $r_{ik}$ and $B_r$ bases, the key projection is
\begin{equation}
\begin{aligned}
 \omega_{ik}^{K}&=\operatorname{softmax}(W_\omega^Kr_{ik}+b_\omega^K),\\
 W_K(r_{ik})&=W_K^{(0)}+
 \sum_{b=1}^{B_r}\omega_{ik,b}^{K}U_{K,b}V_{K,b}^{\top}.
\end{aligned}\label{eq:low_rank_kv}
\end{equation}
Here $U_{K,b},V_{K,b}\in\mathbb R^{d\times r}$ give rank-$r$ updates. The value branch has separate mixture weights and bases, and the token query is $\mathbf q_{ti}^{\ell}=W_Q^{\ell}h_{ti}^{\ell}$. Sharing the bases across paths limits parameters while allowing different relation compositions to change the features used for matching and the evidence passed onward.
At evidence-attention layer $\ell$, token $t$ selects paths using
\begin{equation}
 \alpha_{tik}^{\ell}=\operatorname{softmax}_k\left(
 \frac{(\mathbf q_{ti}^{\ell})^\top k_{ik}^{\ell}}{\sqrt d}
 +\lambda_\ell c_q\log(p_{ik}+\epsilon)\right).
 \label{eq:resemattn_main_revised}
\end{equation}
The dot product supplies language--memory compatibility, while the second term introduces schema compatibility. The resulting context is fused through a learned gate:
\begin{equation}
\begin{aligned}
 c_{ti}^{\ell}&=\sum_k\alpha_{tik}^{\ell}v_{ik}^{\ell},\\
 g_{ti}^{\ell}&=\sigma\!\left(W_g^\ell[h_{ti}^{\ell};c_{ti}^{\ell};h_S;h_{a_i}]\right),\\
 h_{ti}^{\ell+1}&=\operatorname{LN}\!\left(h_{ti}^{\ell}
 +\operatorname{Dropout}(g_{ti}^{\ell}\odot c_{ti}^{\ell})\right).
\end{aligned}\label{eq:gated_fusion_main_revised}
\end{equation}
Here $h_{a_i}$ is the pooled candidate-text embedding; $h_S$ projects the encoding of the serialized schemas and their probabilities. The reference model stacks two such layers after the backbone's returned hidden states. Relation-conditioned projections determine evidence representation; the prior determines relative attention. The gate modulates evidence without guaranteeing its validity.

\noindent\textbf{Matrix computation.}
For a candidate with $K$ retained paths, stack their memories in $Z_i\in\mathbb R^{K\times d}$ and $T$ token states in $H_i\in\mathbb R^{T\times d}$. Relation-conditioned projections produce $K_i,V_i\in\mathbb R^{K\times d}$, and $Q_i=H_iW_Q^{\top}$. The operation in the left panel of Figure~\ref{fig:framework} is
\begin{equation}
\begin{aligned}
 A_i&=\operatorname{softmax}_{\mathrm{row}}\!\left(
 \frac{Q_iK_i^\top}{\sqrt d}+\mathbf1_T\boldsymbol\eta_i^\top\right),\\
 \eta_{ik}&=\lambda c_q\log(p_{ik}+\epsilon),\qquad C_i=A_iV_i.
\end{aligned}\label{eq:matrix_attention}
\end{equation}
The semantic bias is shared across token rows, while the query--key term varies with the token. Thus a common question-level schema preference coexists with token-specific evidence selection. In the imatinib example, the two columns correspond to the target-protein and contraindication paths; the matrices show the computation, not measured attention values. A stronger path prior can favor the former without forcing all tokens to attend identically.

\subsection{Answer Scoring and Absolute Support}\label{subsec:prediction_training_main_revised}
Let $\bar h_i$ denote the mean updated token states. We combine the attended question, candidate, and original question representations with normalized multi-path support:
\begin{equation}
\begin{aligned}
 b_i&=\log\sum_{k=1}^{|\Pi_i|}e^{u_{ik}}-\log|\Pi_i|,\\
 s_i&=\operatorname{MLP}([\bar h_i;h_{a_i};h_q])+\mu b_i,\\
 \pi_i^*&=\arg\max_{\pi_{ik}\in\Pi_i}u_{ik}.
\end{aligned}\label{eq:candidate_score_main_revised}
\end{equation}
We use $b_i$ for scalar candidate support to distinguish it from the value vector $v_{ik}^{\ell}$. Explanations are selected by $u_{ik}$, not by aggregating attention weights.

Normalized attention only compares paths within a candidate. Adding a common constant $b$ to their scores leaves $p_{ik}$ unchanged but changes $b_i$ by $b$. With one path, $p_{i1}=1$ regardless of its quality. The absolute-support term therefore retains information about uniformly weak evidence that the attention normalization removes. Log-mean-exp is invariant to duplicating the entire path multiset, but not to selectively duplicating a strong path; it does not establish evidence independence.

\noindent\textbf{Interpreting a returned path.}
The explanation identifies the most compatible retained path; the candidate score also uses language features and aggregate support. A correct endpoint can therefore score highly while its best path is invalid. Reporting the endpoint, selected path, and independent validity label exposes this failure; attention or within-pool compatibility alone cannot establish support.

\subsection{Training and Output Protocols}\label{subsec:training_outputs}
For candidate labels $y_i=\mathbf1[a_i\in\mathcal A_q]$ and training logits $\widetilde s_i$, the reference objective combines binary cross-entropy, positive-set ranking, and probability calibration:
\begin{equation}
\begin{aligned}
 \mathcal L&=\mathcal L_{\mathrm{BCE}}+0.5\mathcal L_{\mathrm{list}}
                 +0.1\mathcal L_{\mathrm{Brier}},\\
 \mathcal L_{\mathrm{list}}&=\log\sum_i e^{\widetilde s_i}
                       -\log\sum_{i:y_i=1}e^{\widetilde s_i},\\
 \mathcal L_{\mathrm{Brier}}&=|\mathcal C_q|^{-1}
                    \sum_i(\sigma(\widetilde s_i)-y_i)^2.
\end{aligned}\label{eq:training_objective}
\end{equation}
The listwise term is omitted when no positive candidate is retrieved; questions without candidate paths are skipped. Search remains discrete. The loss combines candidate discrimination, ranking over the available candidates, and probability calibration. The scoring and output interfaces use the candidate representation and path support defined above.

\noindent\textbf{Output interfaces.}
Ranking sorts candidates by $s_i$ and attaches $\pi_i^*$ to each returned endpoint. Multi-answer decoding applies a sigmoid-score threshold selected on development macro F1. PrimeKGQA additionally returns directed relation outputs; MedReason maps candidate scores to answer options or normalized answer strings. Appendix Section~\ref{app:decoding} specifies these task interfaces.

\section{Experiments}\label{sec:experiments}
\subsection{Setup and Evaluation}\label{subsec:exp_setup}
\noindent\textbf{Datasets and baselines.}
We study biomedical reasoning using BioHopR, PrimeKGQA,
and MedReason.
The original BioHopR release contains 10,127 unique questions,
including 2,494 one-hop and 7,633 two-hop questions with
multiple reference answers~\citep{kim2025biohopr}.
PrimeKGQA contains 83,999 question--answer pairs accompanied
by SPARQL queries~\citep{primekgqa2024zenodo}.
MedReason contains 32,682 medical question--answer pairs
with step-by-step reasoning explanations~\citep{wu2025medreason}.
These figures describe the released resources.
Task-specific output handling is specified in
Appendix~\ref{app:decoding}.
Baselines cover prompting, retrieval augmentation, biomedical
encoders, graph neural reasoning, and KG exploration
(Table~\ref{tab:main_results_all}).
Controlled variants use fixed candidates, one hard schema,
or the full soft-schema design.

\noindent\textbf{Configuration.}
The reported experiments use Llama-3.1-8B-Instruct and five seeds on a workstation with four 24-GB RTX~4090 GPUs. The search setting uses $N=8$, $M=4$, $B=32$, at most three hops, and $K=8$. Appendix Section~C gives the model configuration and seed protocol.

\noindent\textbf{Answer and evidence metrics.}
BioHopR uses H@1/5/10, P@10, R@10, and MRR; PrimeKGQA uses answer-set EM/F1/recall and relation F1; MedReason uses multiple-choice accuracy/MRR and open-answer EM/token F1. Coverage requires at least one gold entity in the retrieved pool; candidate recall averages the fraction of each gold answer set retrieved. On a common annotated cohort, let $A_q$ denote top-answer correctness and $V_q$ the validity of its selected supporting path. Then
\begin{equation}
\begin{aligned}
 \mathrm{ValidPath@1}&=\mathbb E_q[V_q],\qquad
 \mathrm{Joint@1}=\mathbb E_q[A_qV_q],\\
 \mathrm{WPR}&=\frac{\sum_q A_q(1-V_q)}{\sum_q A_q}.
\end{aligned}\label{eq:evidence_metrics}
\end{equation}
A missing path has $V_q=0$. WPR is undefined when no top-ranked answer is correct. For each run, H@1, ValidPath@1, Joint@1, and WPR use the same evaluation questions and path-validity labels; consequently, $\mathrm{Joint@1}=\mathrm{H@1}(1-\mathrm{WPR})$. Appendix Section~\ref{app:metrics} gives the ranking and aggregation conventions.

\begin{table*}[!t]
\centering
\small
\setlength{\tabcolsep}{1.75pt}
\begin{adjustbox}{max width=\linewidth}
\begin{tabular}{@{}lccccccccccc@{}}
\toprule
& \multicolumn{6}{c}{\textbf{\biohopr{}}}
& \multicolumn{3}{c}{\textbf{PrimeKGQA}}
& \multicolumn{2}{c}{\textbf{MedReason}} \\
\cmidrule(lr){2-7}\cmidrule(lr){8-10}\cmidrule(lr){11-12}
\textbf{Method}
& \textbf{H@1} & \textbf{H@5} & \textbf{H@10}
& \textbf{P@10} & \textbf{R@10} & \textbf{MRR}
& \textbf{EM} & \textbf{F1} & \textbf{Rec.}
& \textbf{MCQ Acc.} & \textbf{Open F1} \\
\midrule
Zero-shot
& 0.221 & 0.352 & 0.411 & 0.176 & 0.286 & 0.294
& 0.342 & 0.487 & 0.522 & 0.512 & 0.538 \\
CoT
& 0.468 & 0.474 & 0.515 & 0.313 & 0.367 & 0.486
& 0.368 & 0.516 & 0.548 & 0.548 & 0.566 \\
Dense RAG
& 0.403 & 0.558 & 0.606 & 0.337 & 0.421 & 0.471
& 0.394 & 0.543 & 0.579 & 0.563 & 0.581 \\
GraphRAG
& 0.548 & 0.641 & 0.674 & 0.402 & 0.478 & 0.601
& 0.427 & 0.574 & 0.612 & 0.586 & 0.604 \\
GNN-RAG
& 0.587 & 0.672 & 0.708 & 0.425 & 0.501 & 0.635
& 0.444 & 0.591 & 0.629 & 0.604 & 0.621 \\
BioLinkBERT
& 0.579 & 0.661 & 0.697 & 0.420 & 0.491 & 0.624
& 0.443 & 0.589 & 0.627 & 0.581 & 0.599 \\
PullNet
& 0.574 & 0.604 & 0.631 & 0.414 & 0.493 & 0.596
& 0.451 & 0.596 & 0.632 & 0.591 & 0.609 \\
GreaseLM
& 0.626 & 0.702 & 0.745 & 0.451 & 0.522 & 0.684
& 0.476 & 0.617 & 0.653 & 0.611 & 0.629 \\
UniKGQA
& 0.643 & 0.709 & 0.748 & 0.466 & 0.536 & 0.696
& 0.488 & 0.628 & 0.664 & 0.620 & 0.638 \\
Think-on-Graph
& 0.655 & 0.718 & 0.756 & 0.472 & 0.542 & 0.704
& 0.495 & 0.634 & 0.671 & 0.626 & 0.644 \\
\midrule
Fixed-Candidate \coremethod{}
& 0.719 & 0.736 & 0.743 & 0.582 & 0.556 & 0.726
& 0.508 & 0.642 & 0.678 & 0.637 & 0.655 \\
\framework{} w/ Hard Schema
& 0.748 & 0.789 & 0.811 & 0.598 & 0.621 & 0.765
& 0.531 & 0.661 & 0.696 & 0.652 & 0.670 \\
\textbf{\framework{} w/ Soft Schema}
& \textbf{0.772} & \textbf{0.816} & \textbf{0.842}
& \textbf{0.617} & \textbf{0.648} & \textbf{0.790}
& \textbf{0.652} & \textbf{0.761} & \textbf{0.794}
& \textbf{0.691} & \textbf{0.749} \\
\bottomrule
\end{tabular}
\end{adjustbox}
\caption{Main results across three datasets. Bold denotes the best reported value in each column.}
\label{tab:main_results_all}
\end{table*}
\begin{table*}[!t]
\centering
\small
\setlength{\tabcolsep}{2.25pt}
\begin{adjustbox}{max width=\linewidth}
\begin{tabular}{@{}lccccccc@{}}
\toprule
& \multicolumn{2}{c}{\textbf{\biohopr{}}}
& \multicolumn{3}{c}{\textbf{PrimeKGQA}}
& \multicolumn{2}{c}{\textbf{MedReason}} \\
\cmidrule(lr){2-3}\cmidrule(lr){4-6}\cmidrule(lr){7-8}
\textbf{Variant}
& \textbf{H@1} & \textbf{MRR}
& \textbf{Ans. EM} & \textbf{Ans. F1} & \textbf{Rel. F1}
& \textbf{MCQ Acc.} & \textbf{Open F1} \\
\midrule
w/o Grounding Distribution
& 0.742 & 0.764 & 0.521 & 0.651 & 0.708 & 0.642 & 0.659 \\
w/o Relation-Guided Search
& 0.719 & 0.744 & 0.503 & 0.633 & 0.684 & 0.631 & 0.648 \\
w/ Hard Question Schema
& 0.748 & 0.765 & 0.531 & 0.661 & 0.714 & 0.652 & 0.670 \\
w/o Schema Entropy Gate
& 0.758 & 0.776 & 0.541 & 0.671 & 0.729 & 0.661 & 0.680 \\
w/o Attention-Prior Injection
& 0.706 & 0.739 & 0.494 & 0.625 & 0.675 & 0.620 & 0.641 \\
w/o Relation-Conditioned KV
& 0.731 & 0.756 & 0.514 & 0.644 & 0.694 & 0.638 & 0.657 \\
w/o Candidate Path Aggregation
& 0.745 & 0.768 & 0.526 & 0.655 & 0.706 & 0.648 & 0.666 \\
\textbf{Full \framework{}}
& \textbf{0.772} & \textbf{0.790}
& \textbf{0.652} & \textbf{0.761} & \textbf{0.770}
& \textbf{0.691} & \textbf{0.749} \\
\bottomrule
\end{tabular}
\end{adjustbox}
\caption{Core ablation across the three datasets.}
\label{tab:cross_dataset_ablation}
\end{table*}
\subsection{Answer Quality Across Datasets}\label{subsec:main_results}
Table~\ref{tab:main_results_all} reports the strongest answer performance for the soft-schema model across all listed metrics. On BioHopR, it improves H@1 from 0.719 to 0.772 and R@10 from 0.556 to 0.648 over Fixed-Candidate \coremethod{}. The larger recall gain is consistent with improved candidate discovery. PrimeKGQA answer F1 rises from 0.661 with a hard schema to 0.761 with soft schemas; MedReason reaches 0.691 multiple-choice accuracy and 0.749 open-answer F1.
BioHopR emphasizes recovery of multiple gold entities, PrimeKGQA evaluates answer sets and relations, and MedReason separates option selection from open-answer overlap. Gains across these formats support relation-semantic evidence; absolute metric values are not comparable across datasets. Appendix Section~\ref{app:dataset_results} retains all precision, recall, relation-F1, and open-answer metrics. Baseline configurations and adaptations are described in Appendix~\ref{app:baseline_protocol}.

\noindent\textbf{Variation across seeds.}
The five-seed BioHopR H@1 means and standard deviations are $0.772\pm0.006$ for Full, $0.748\pm0.007$ for Hard-Schema, and $0.719\pm0.008$ for Fixed-Candidate. The corresponding Joint@1 values are $0.602\pm0.007$, $0.548\pm0.009$, and $0.495\pm0.010$. These summaries describe run variation; aggregate means alone do not establish paired significance. Appendix Section~\ref{app:statistics} gives the seed-summary protocol.

\subsection{Core Components Across Datasets}\label{subsec:cross_dataset_ablation}
Table~\ref{tab:cross_dataset_ablation} separates the core components. Removing relation-guided search lowers BioHopR H@1 by 5.3 percentage points; removing the attention prior lowers it by 6.6 points, the largest decrease among these core ablations. Both relation-conditioned projections and soft schemas contribute across datasets. Aggregation also improves answer-set quality, consistent with combining evidence for multiple answers.

Grounding and search determine available evidence; projections and the prior determine how it enters the representation. Hard schemas affect both stages. Because end-to-end ablations may change path pools, they show component sensitivity; frozen-pool comparisons isolate effects on identical evidence.

\begin{figure*}[!t]
\centering
\begin{minipage}[t]{0.138\textwidth}\centering
\includegraphics[width=\linewidth]{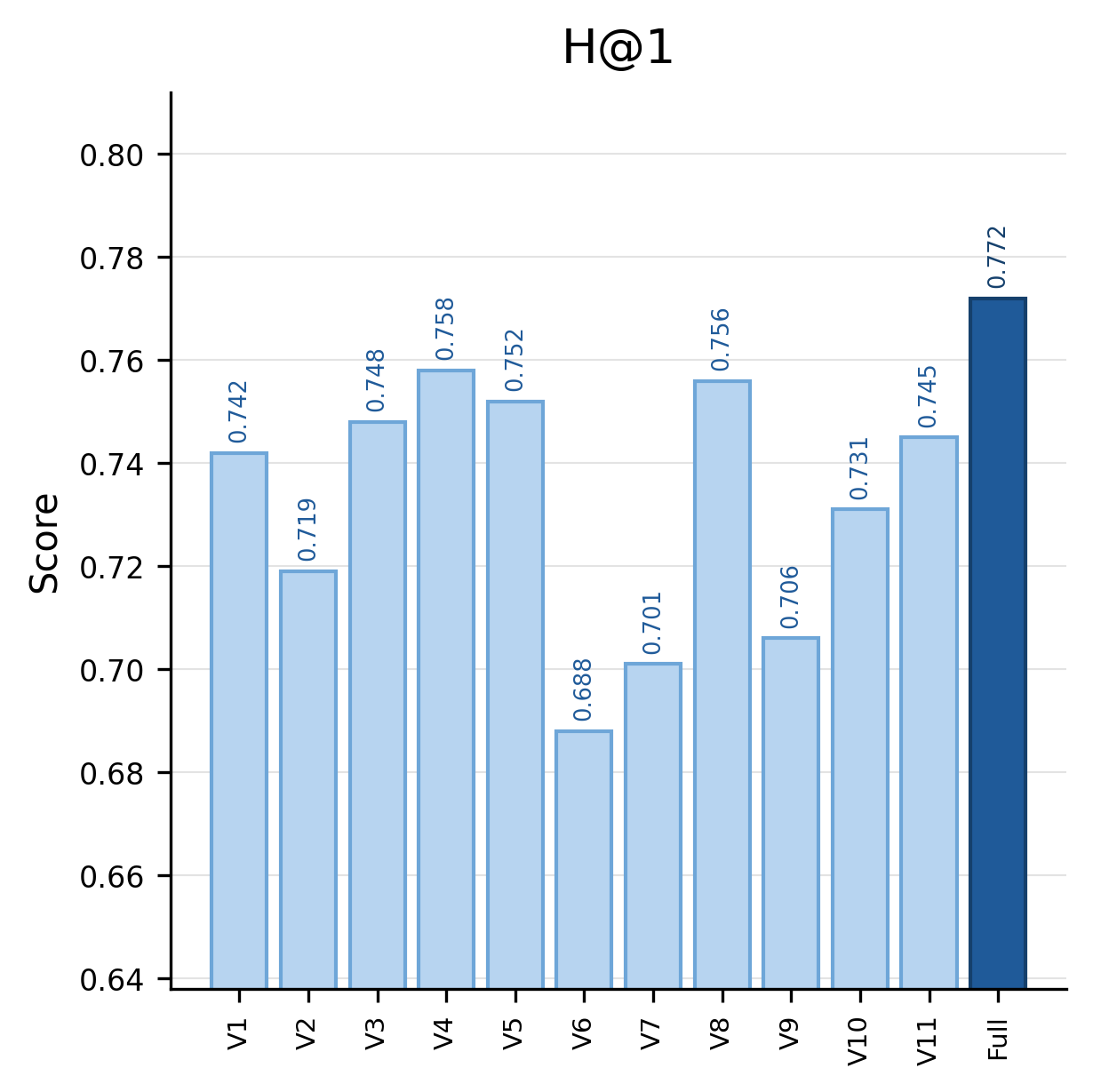}
\end{minipage}\hfill
\begin{minipage}[t]{0.138\textwidth}\centering
\includegraphics[width=\linewidth]{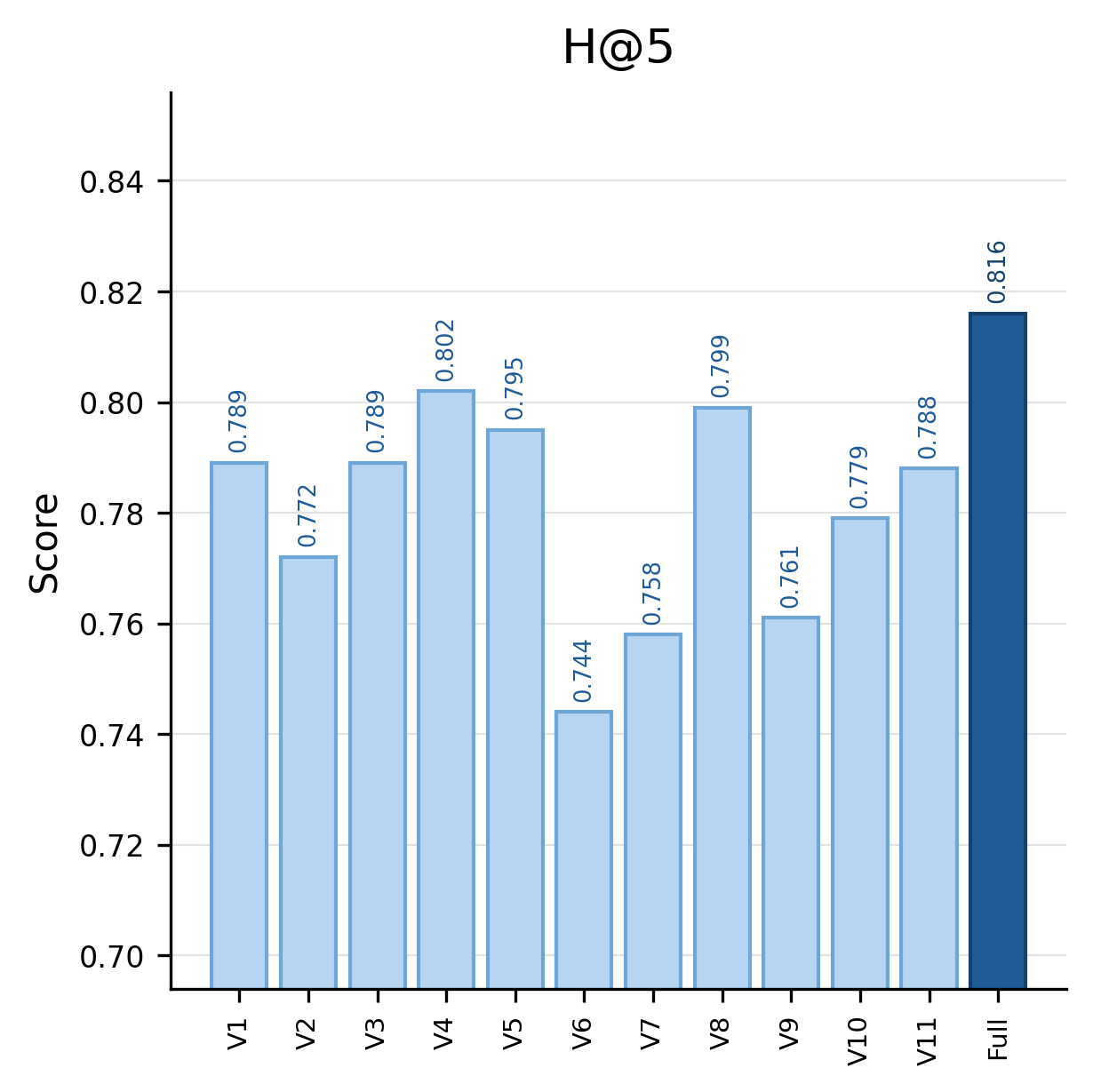}
\end{minipage}\hfill
\begin{minipage}[t]{0.138\textwidth}\centering
\includegraphics[width=\linewidth]{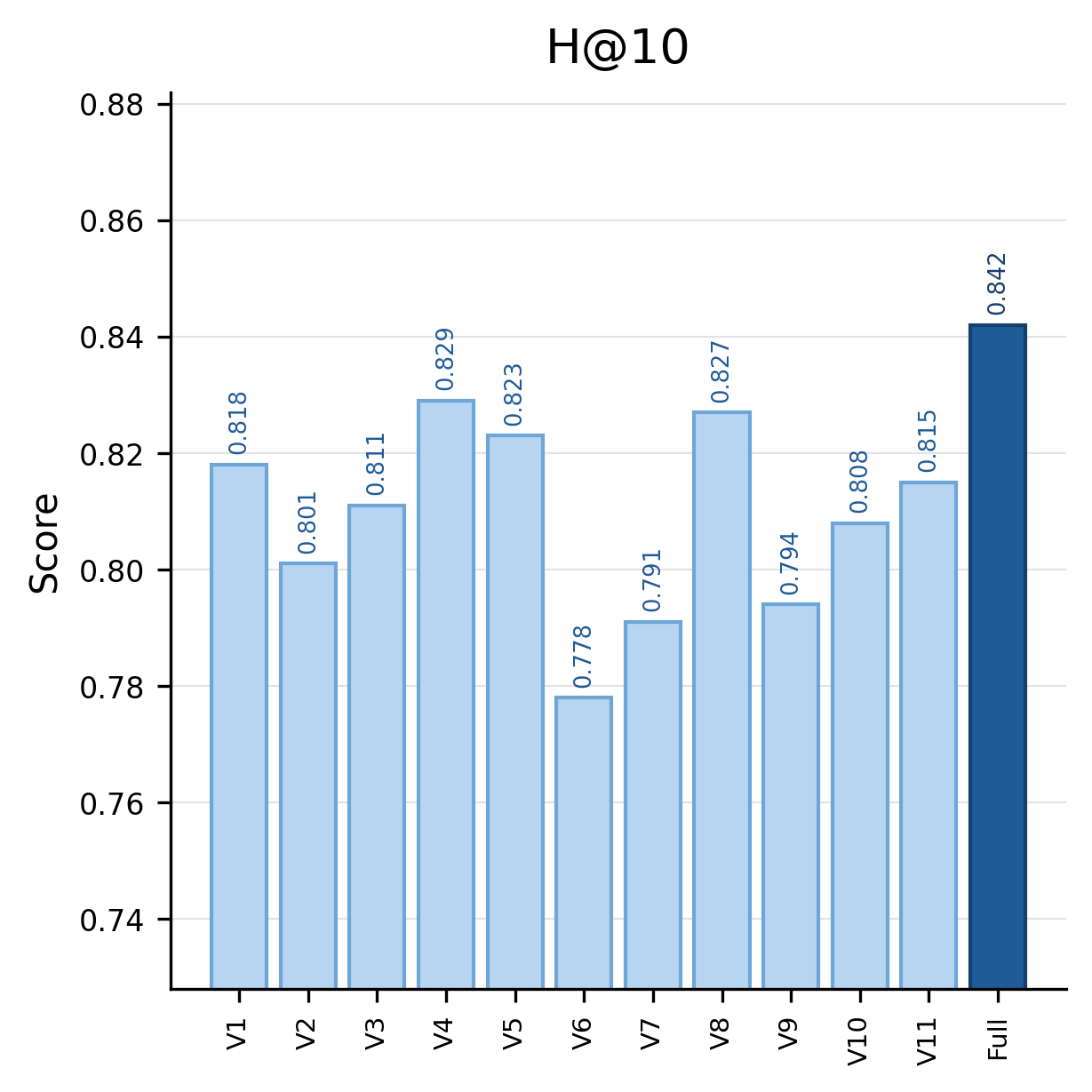}
\end{minipage}\hfill
\begin{minipage}[t]{0.138\textwidth}\centering
\includegraphics[width=\linewidth]{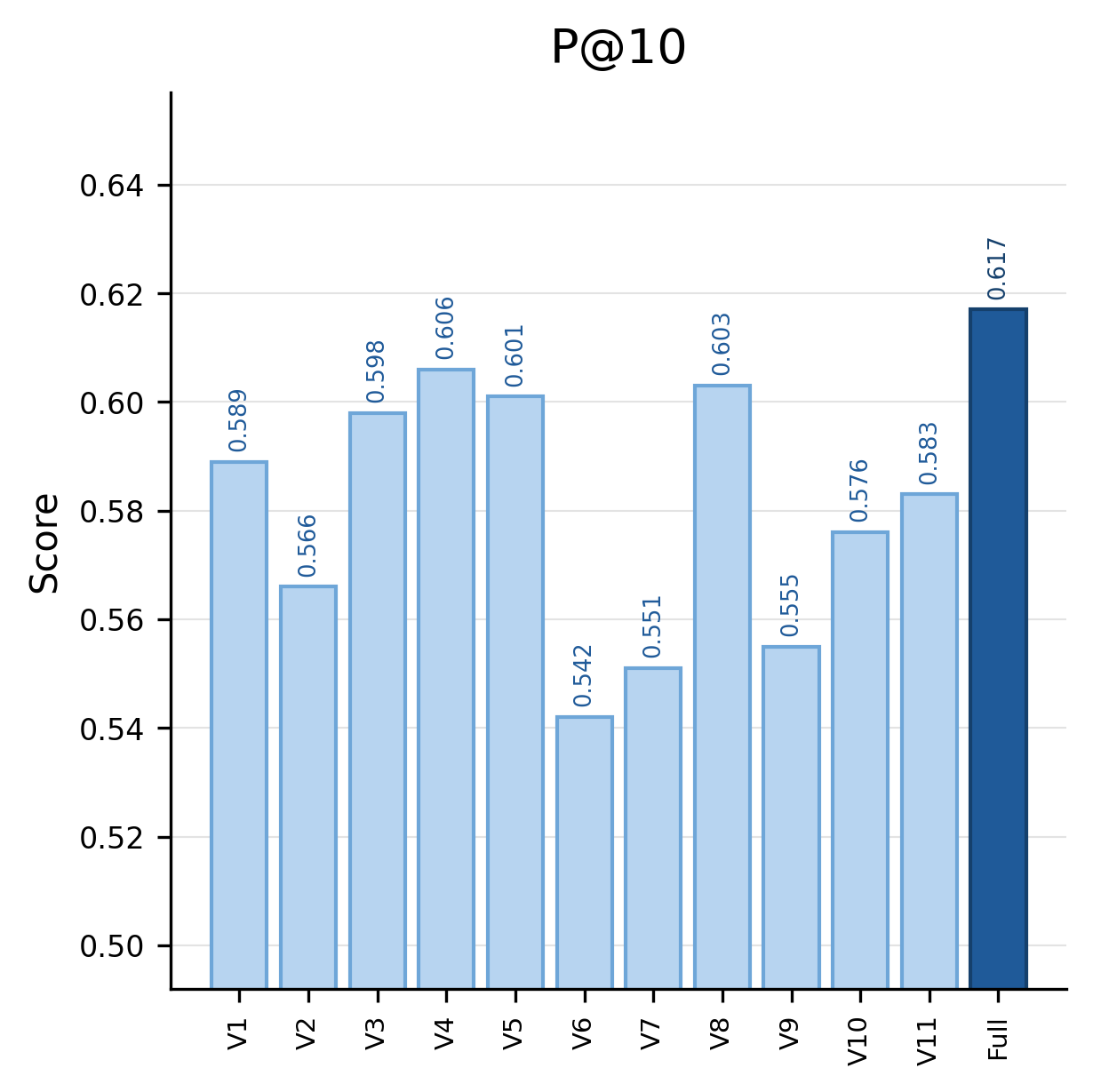}
\end{minipage}\hfill
\begin{minipage}[t]{0.138\textwidth}\centering
\includegraphics[width=\linewidth]{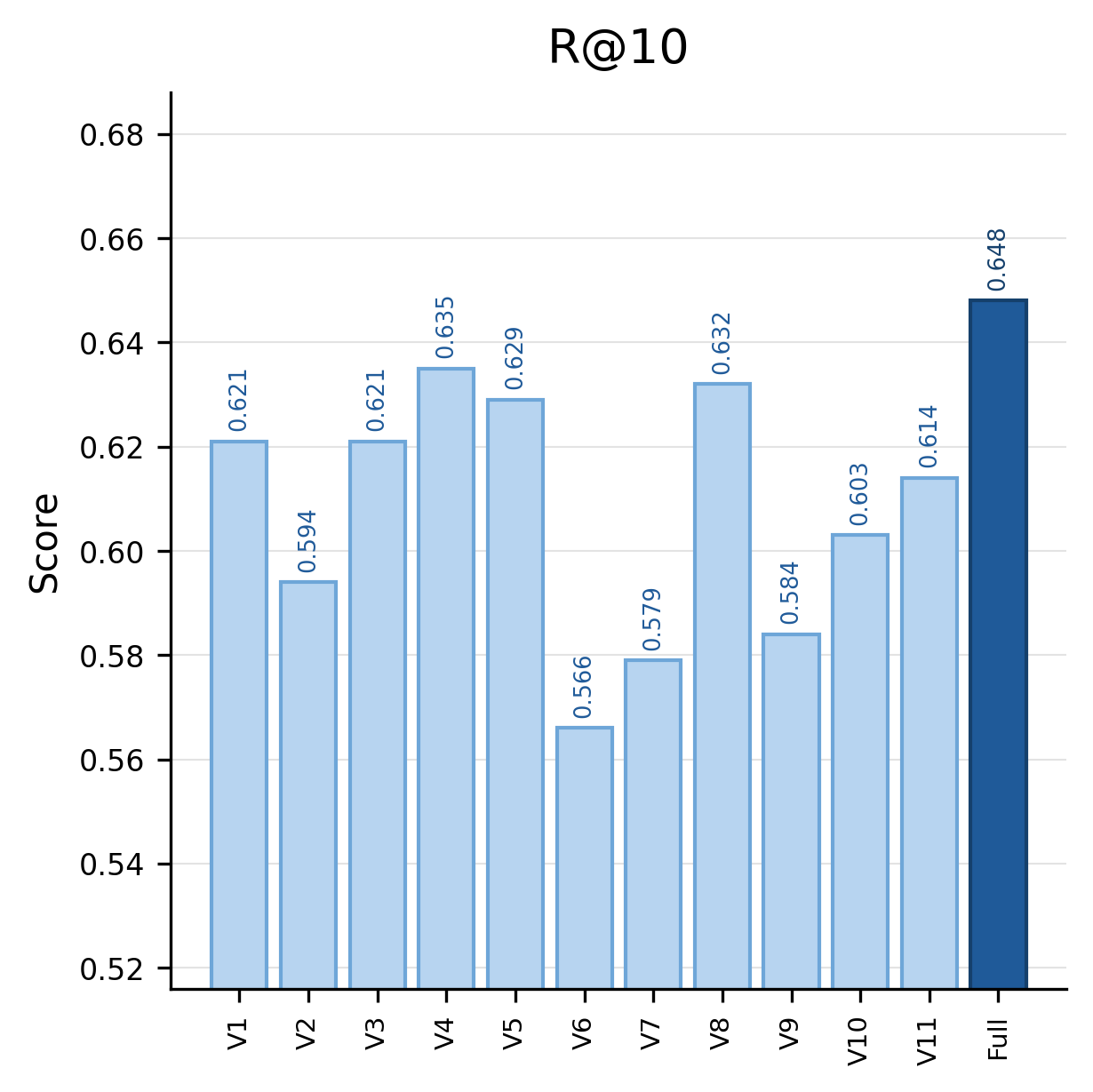}
\end{minipage}\hfill
\begin{minipage}[t]{0.138\textwidth}\centering
\includegraphics[width=\linewidth]{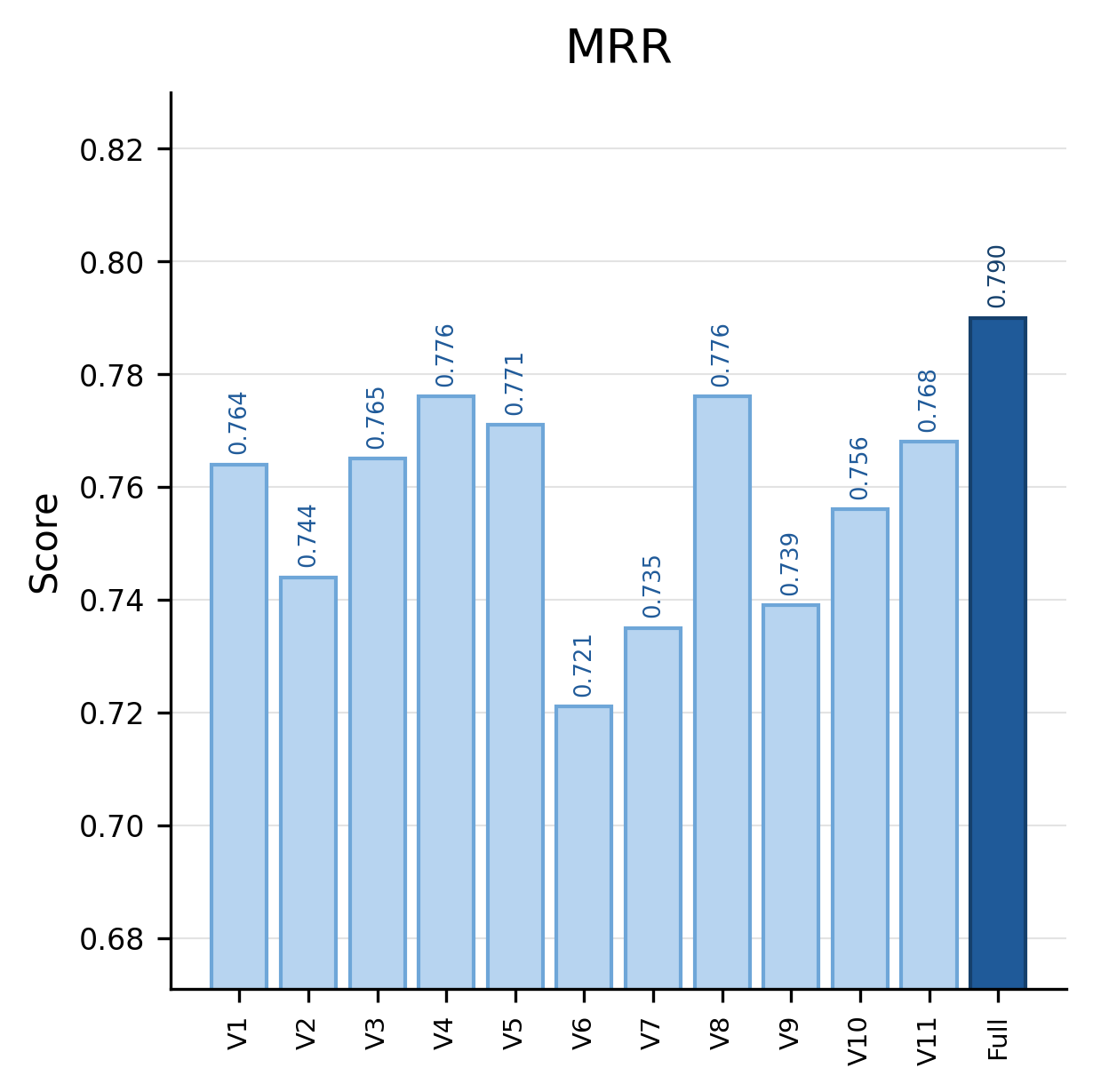}
\end{minipage}\hfill
\begin{minipage}[t]{0.138\textwidth}\centering
\includegraphics[width=\linewidth]{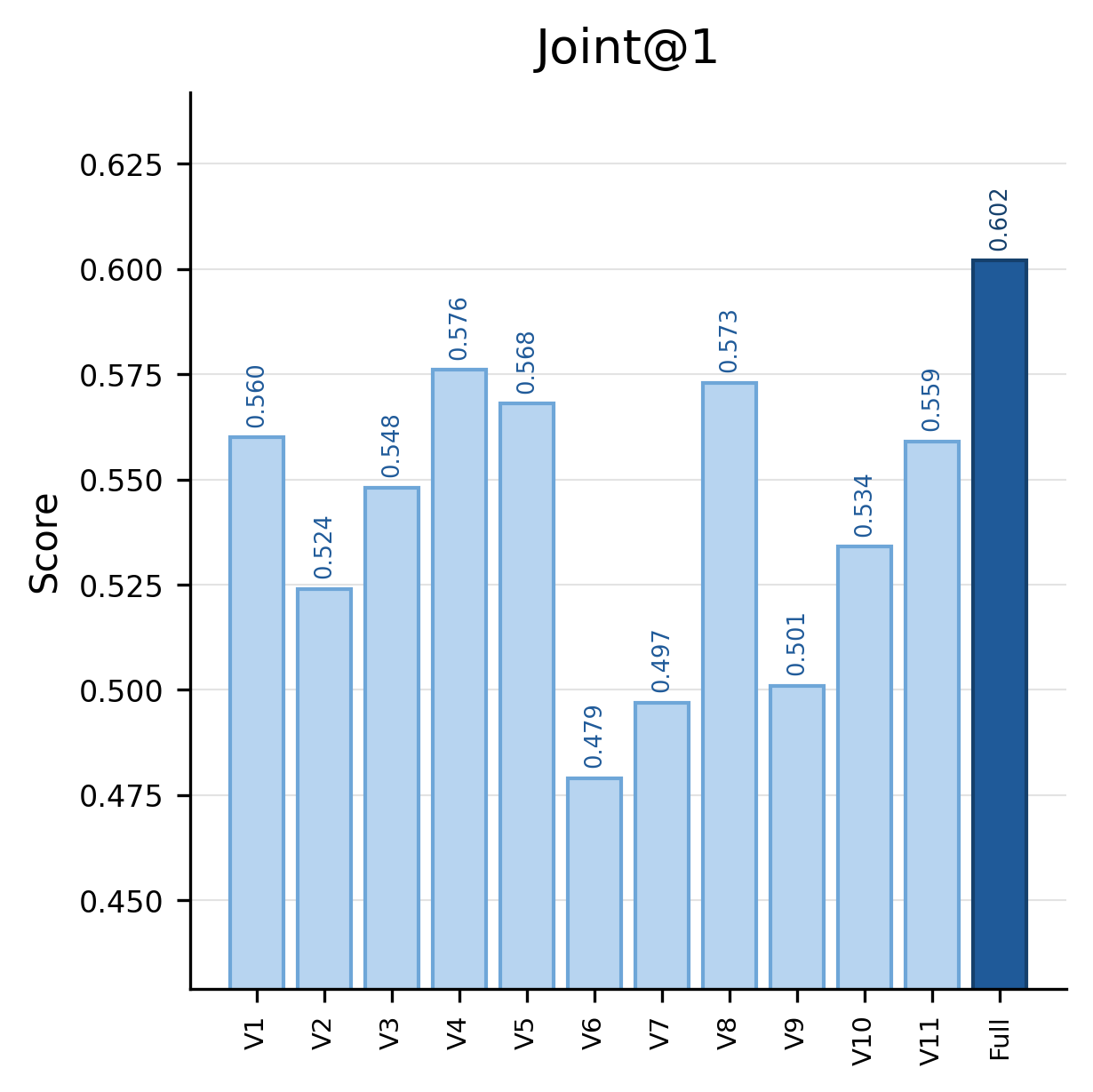}
\end{minipage}
\caption{Per-metric ablation results on \biohopr{}. V1--V11 denote the eleven ablated variants, and Full denotes the complete \framework{} model.
    V1  = w/o Grounding Distribution
V2  = w/o Relation-Guided Search
V3  = w/ Hard Question Schema
V4  = w/o Schema Entropy Gate
V5  = w/o Entity Anchoring
V6  = w/o Relation Compatibility
V7  = w/o Path Consistency
V8  = w/o Topology Penalty
V9  = w/o Attention Prior
V10 = w/o Relation-Conditioned KV
V11 = w/o Path Aggregation. Higher is better for all metrics.}
\label{fig:biohopr_ablation_bars_blue}
\end{figure*}
\subsection{Detailed BioHopR Ablation}\label{subsec:biohopr_ablation}
Figure~\ref{fig:biohopr_ablation_bars_blue} reports all seven metrics: H@1/5/10 assess answer ranking, P@10 and R@10 assess answer-set recovery, MRR emphasizes the first correct answer, and Joint@1 additionally requires valid evidence. Their joint interpretation separates endpoint ranking from evidence selection.

\noindent\textbf{Availability and evidence factors.}
V1--V4 vary grounding, search, and schema uncertainty; V5--V8 remove schema-matching factors. Declines without relation compatibility or path consistency show that endpoint type is insufficient. V9--V10 isolate the attention prior and relation-conditioned projections within the full pipeline. These end-to-end changes complement, rather than replace, matched-pool controls.

\noindent\textbf{Combining paths.}
V11 removes aggregation. The aggregation comparison reports H@1/R@10 of 0.772/0.648 for normalized LogSumExp, 0.754/0.619 for max pooling, and 0.739/0.607 for mean pooling. Normalized aggregation improves answer-set recovery here without establishing independence of paths sharing edges. Appendix Section~E.3 retains all variants, including unnormalized LogSumExp and learned gated pooling.

\subsection{Correct Endpoints and Valid Evidence}\label{subsec:beyond_reachability}\label{subsec:running_case_results}
The reachable-but-invalid challenge compares real KG paths containing relation swaps, direction errors, bridge confounders, or right-answer/wrong-path alternatives. Paths are matched by hop count, endpoint type, degree bucket, and initial retrieval-score bucket. Blinded annotation judges the complete question--path relation, rather than agreement with the model's predicted schema; construction and agreement measurements are retained in Appendix Section~D.

Two annotators label each path Valid, Invalid, or Insufficient/Ambiguous; a third adjudicates disagreements. They see the question, candidate, directed path, relation definitions, and available provenance, with model identity, scores, and predicted schemas hidden. Reported agreement is 0.884 overall ($\kappa=0.761$) and 0.901 on correct-answer paths ($\kappa=0.792$). The latter subset directly tests whether endpoint correctness masks disagreement about the evidence. A legal inverse traversal is not itself an invalid direction reversal.

In the running imatinib case, the two paths share the CML endpoint but express different relation compositions: the ABL1 route follows the requested target-protein relation, whereas the contraindication route does not.

\begin{table*}[!t]
\centering\small
\setlength{\tabcolsep}{4pt}
\textbf{E1. Frozen candidate/path pools}\hfill\\[3pt]
\begin{tabular*}{\textwidth}{@{\extracolsep{\fill}}llrrrrrr@{}}
\toprule
Retriever & Reasoner & Coverage & CandRec & H@1 & Cond. H@1 & ValidPath@1 & Joint@1\\
\midrule
Unguided & CE & 0.860&0.691&0.639&0.743&0.520&0.462\\
Unguided & ReSemAttn & 0.860&0.691&0.701&0.815&0.646&0.597\\
Relation-guided & CE & 0.947&0.792&0.692&0.731&0.585&0.541\\
Relation-guided & ReSemAttn & 0.947&0.792&0.766&0.809&0.714&0.681\\
\bottomrule
\end{tabular*}
\par\medskip
\begin{minipage}[t]{0.49\textwidth}
\centering\textbf{E2. Relation integration}\hfill\\[3pt]
\begin{tabularx}{\linewidth}{@{}Yrrr@{}}
\toprule
Integration & H@1 & PairAcc & Joint@1\\
\midrule
Labels in text & 0.696&0.625&0.545\\
Final-score only & 0.718&0.696&0.579\\
Prior + shared KV & 0.744&0.763&0.634\\
Prior + relation KV & 0.766&0.819&0.681\\
\bottomrule
\end{tabularx}
\end{minipage}\hfill
\begin{minipage}[t]{0.49\textwidth}
\centering\textbf{E5. Absolute support}\hfill\\[3pt]
\begin{tabularx}{\linewidth}{@{}Yrrr@{}}
\toprule
Variant & \shortstack{Normal\\H@1} & \shortstack{Invalid\\FSR$\downarrow$} & \shortstack{One-path\\FSR$\downarrow$}\\
\midrule
Without $\mu b_i$ & 0.758&0.407&0.456\\
Full & 0.766&0.169&0.203\\
\bottomrule
\end{tabularx}
\end{minipage}
\caption{Consolidated mechanism controls. Cond.\ H@1 conditions on a retrieved gold answer; PairAcc compares same-answer valid/invalid paths. FSR is acceptance of an all-invalid candidate pool; the one-path condition also uses invalid evidence.}
\label{tab:mechanism_controls}\label{tab:pending_e1}\label{tab:pending_e2}\label{tab:pending_e5}
\end{table*}

\begin{figure}[!htbp]
    \centering
    \includegraphics[width=\columnwidth]
    {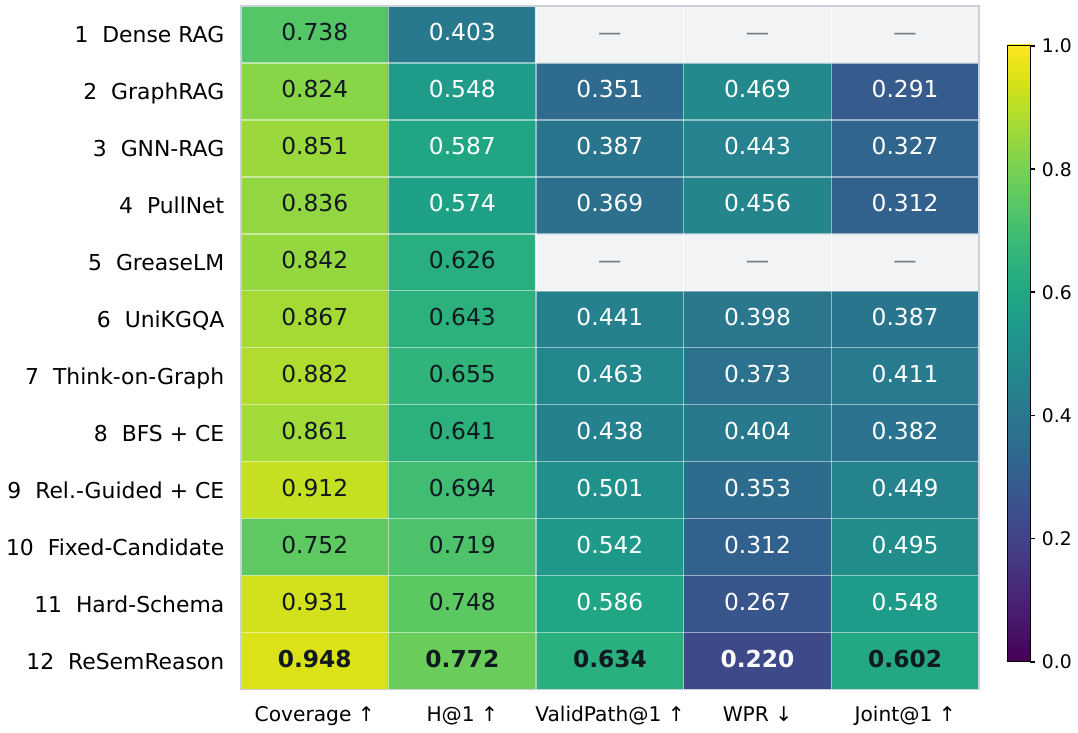}
    \caption{Answer availability and evidence quality. Rows 1--12 identify methods; CE denotes Cross-Encoder. Higher is better except for WPR. A dash denotes an unreported path metric.}
\label{fig:reachable_invalid_heatmap}
\end{figure}

Figure~\ref{fig:reachable_invalid_heatmap} separates availability from evidence quality. With the same Cross-Encoder, relation-guided search raises Coverage from 0.861 to 0.912 and H@1 from 0.641 to 0.694. Fixed-Candidate \coremethod{} reaches H@1 of 0.719 with Coverage of 0.752, illustrating how strong ranking can coexist with a limited candidate pool. The full model reports Coverage 0.948, H@1 0.772, ValidPath@1 0.634, and Joint@1 0.602. Because the compared systems do not share identical pools, these results motivate the frozen-pool control rather than replacing it.

\begin{figure}[!htbp]
    \centering
    \begin{minipage}[t]{0.49\columnwidth}
        \centering
        \includegraphics[width=\linewidth]{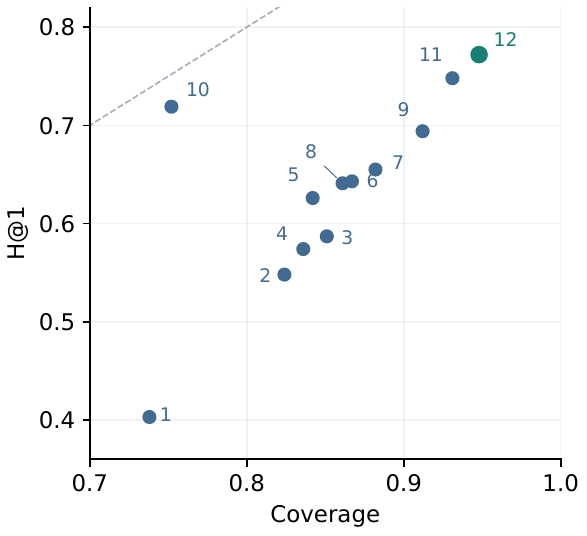}
    \end{minipage}\hfill
    \begin{minipage}[t]{0.49\columnwidth}
        \centering
        \includegraphics[width=\linewidth]{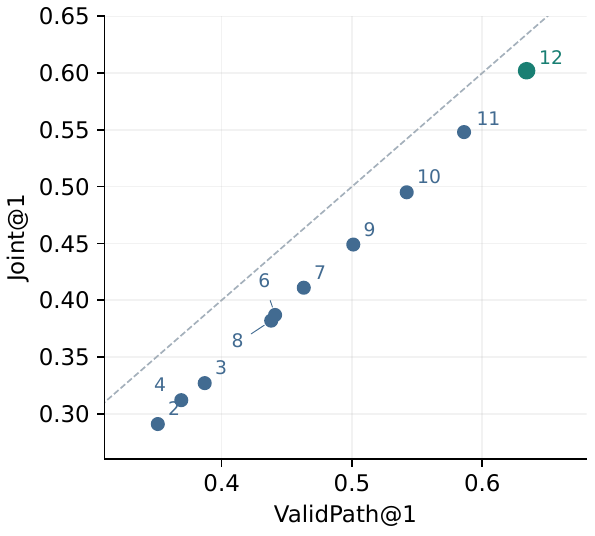}
    \end{minipage}
    \caption{Candidate availability and evidence quality. Numeric method IDs follow Figure~\ref{fig:reachable_invalid_heatmap}; methods 1 (Dense RAG) and 5 (GreaseLM) do not return paths and are omitted on the right.}
\label{fig:reachable_invalid_two_views_square}
\end{figure}
Figure~\ref{fig:reachable_invalid_two_views_square} separates candidate discovery from ranking (Coverage--H@1), then incorporates evidence validity (ValidPath@1--Joint@1). Method IDs match the heatmap. Path-free methods remain in the answer comparison but are omitted from path metrics; missing values are not zeros.

On same-answer valid/invalid pairs, PairAcc requires a strictly higher within-model score for the valid path; ties receive no credit. Reported values are 0.493 for BFS + CE, 0.658 for Fixed-Candidate \coremethod{}, 0.708 for Hard-Schema, and 0.806 for Full. Holding the gold endpoint fixed isolates path preference. Appendix Section~D describes the negative-path construction.

\noindent\textbf{Where evidence remains difficult.}
Among 125 manually categorized failures, 27 involve search pruning, 23 a missing top-$M$ schema, and 19 grounding. These three categories account for 55.2\% of the analyzed failures and locate bottlenecks in candidate discovery. Appendix Section~\ref{app:case_studies} gives the complete category counts.

\noindent\textbf{Robustness beyond the standard questions.}
Template-disjoint H@1/Joint@1 is 0.741/0.563, versus 0.772/0.602 on the standard split; combined paraphrase and alias perturbations yield 0.731/0.548. These results expose sensitivity to wording and grounding. Template separation does not by itself establish generalization to unseen relation compositions.

With 10\% type-compatible distractors, Coverage/H@1/Joint@1 falls from 0.948/0.772/0.602 to 0.931/0.748/0.567, consistent with stronger evidence competition as well as candidate loss. Removing 10\% of non-gold edges yields 0.934/0.754/0.579; it does not test loss of required gold evidence.

\subsection{Mechanism Controls}\label{subsec:additional_controls}
The existing prior-by-KV factorial experiment reports H@1/\allowbreak MRR/\allowbreak Joint@1 of 0.772/0.790/0.602 with both mechanisms and 0.681/0.716/0.462 with neither; all four settings remain in Appendix Section~E.2. This pattern supports complementary contributions but does not establish a statistical interaction.

Schema prediction is imperfect: reported accuracy is 0.912 for hop count, 0.847 for relation sequences, and 0.781 for full typed schemas, with expected calibration error 0.028, 0.041, and 0.057. These results motivate preserving alternatives and evaluating confidence separately from top-$M$ coverage.

Table~\ref{tab:mechanism_controls} presents values for three matched-control designs. E1 freezes each retriever\textquotesingle s candidate/path pool before either reasoner is applied, so Coverage and candidate recall are shared within a retriever block. E2 fixes relation-guided pools and varies where relation information enters, using same-answer valid/invalid pairs for PairAcc. E5 removes only the absolute-support term and evaluates both normal and all-invalid pools, including a one-path condition. These controls separate retrieval availability, relation-conditioned representation, and absolute support.

\subsection{Efficiency}\label{subsec:efficiency}
Profiling uses one RTX~4090, batch size one, 50 warm-up and 500 timed queries, excluding loading. Full \framework{} uses 276 ms/query and 760 expanded nodes (Table~\ref{tab:end_to_end_efficiency}). Appendix Section~\ref{app:complexity} relates the measured end-to-end latency to the computational structure.

\begin{table}[!htbp]
\centering
\small
\setlength{\tabcolsep}{2.0pt}
\begin{adjustbox}{max width=\linewidth}
\begin{tabular}{@{}lrrrr@{}}
\toprule
\textbf{Method}
& \shortstack{\textbf{Mean}\\\textbf{ms/q}$\downarrow$}
& \shortstack{\textbf{Peak}\\\textbf{GB}$\downarrow$}
& \shortstack{\textbf{Expanded}\\\textbf{Nodes}$\downarrow$}
& \textbf{H@1}$\uparrow$ \\
\midrule
Dense RAG          & 118 & 8.6  & --          & 0.403 \\
GraphRAG           & 242 & 10.1 & 1280 & 0.548 \\
GNN-RAG            & 286 & 11.8 & 1460 & 0.587 \\
PullNet            & 218 & 9.7  & 1320 & 0.574 \\
GreaseLM           & 171 & 12.5 & --          & 0.626 \\
UniKGQA            & 205 & 10.9 & 980  & 0.643 \\
Think-on-Graph     & 514 & 14.2 & 1730 & 0.655 \\
\textbf{Full \framework{}}
                   & \textbf{276}
                   & \textbf{19.8}
                   & \textbf{760}
                   & \textbf{0.772} \\
\bottomrule
\end{tabular}
\end{adjustbox}
\caption{End-to-end efficiency on \biohopr{}. All systems are evaluated with batch size one on identical hardware. A dash indicates that expanded-node counts are not directly applicable or exposed.}
\label{tab:end_to_end_efficiency}
\end{table}

For $C$ candidates, $K$ paths each, $T$ tokens, and hidden width $d$, token-to-path attention costs $O(CTKd)$ per layer, excluding backbone encoding and projections. Appendix Section~\ref{app:complexity} accounts for these costs and distinguishes edge attempts from unique-node expansions.

Increasing beam width from 8 to 32 raises Coverage from 0.883 to 0.948 and H@1 from 0.721 to 0.772, while latency rises from 188 to 276 ms. Beam 64 gives Coverage 0.956 and H@1 0.776 at 391 ms, showing diminishing returns from additional search in this configuration.

\section{Limitations}\label{sec:limitations}
Missing or ambiguous KG relations, incomplete grounding, omitted schemas, and beam pruning can prevent valid evidence from reaching the reasoner. Mean-pooled relation features do not themselves preserve relation order, and the topology penalty is not a provenance-quality model. Retrieved paths and attention weights are inspectable model evidence, not causal explanations or clinical validation.

\section{Conclusion}\label{sec:conclusion}
\framework{} makes relation composition explicit by sharing question-induced schemas between candidate discovery and path attention. Relation-conditioned projections encode evidence, the prior steers selection, and absolute support preserves information lost by normalization. Reported answer, ablation, and path-discrimination results motivate evaluating availability and validity separately. The separation between candidate availability, answer correctness, and supporting-path validity provides a basis for evaluating biomedical multi-hop reasoning.

\clearpage
\bibliographystyle{ACM-Reference-Format}
\bibliography{ReSemAttn_refs}

@inproceedings{xu2025biomedkg,
  title     = {A Survey on Unifying Large Language Models and
               Knowledge Graphs for Biomedicine and Healthcare},
  author    = {Xu, Ran and
               Jiang, Patrick and
               Luo, Linhao and
               Xiao, Cao and
               Cross, Adam and
               Pan, Shirui and
               Sun, Jimeng and
               Yang, Carl},
  booktitle = {Proceedings of the 31st ACM SIGKDD Conference
               on Knowledge Discovery and Data Mining V.2},
  pages     = {6195--6205},
  year      = {2025},
  publisher = {Association for Computing Machinery},
  doi       = {10.1145/3711896.3736556},
  url       = {https://doi.org/10.1145/3711896.3736556}
}

@inproceedings{li2025ragreasoning,
  title     = {A Survey of {RAG}-Reasoning Systems
               in Large Language Models},
  author    = {Li, Yangning and
               Zhang, Weizhi and
               Yang, Yuyao and
               Huang, Wei-Chieh and
               Wu, Yaozu and
               Luo, Junyu and
               Bei, Yuanchen and
               Zou, Henry Peng and
               Luo, Xiao and
               Zhao, Yusheng and
               Chan, Chunkit and
               Chen, Yankai and
               Deng, Zhongfen and
               Li, Yinghui and
               Zheng, Hai-Tao and
               Li, Dongyuan and
               Jiang, Renhe and
               Zhang, Ming and
               Song, Yangqiu and
               Yu, Philip S.},
  booktitle = {Findings of the Association for Computational
               Linguistics: EMNLP 2025},
  pages     = {12120--12145},
  year      = {2025},
  publisher = {Association for Computational Linguistics},
  doi       = {10.18653/v1/2025.findings-emnlp.648},
  url       = {https://aclanthology.org/2025.findings-emnlp.648/}
}

@inproceedings{luo2024rog,
  title     = {Reasoning on Graphs: Faithful and Interpretable
               Large Language Model Reasoning},
  author    = {Luo, Linhao and
               Li, Yuan-Fang and
               Haffari, Reza and
               Pan, Shirui},
  booktitle = {The Twelfth International Conference
               on Learning Representations},
  year      = {2024},
  url       = {https://openreview.net/forum?id=ZGNWW7xZ6Q}
}

@inproceedings{mavromatis2025gnnrag,
  title     = {{GNN-RAG}: Graph Neural Retrieval for Efficient
               Large Language Model Reasoning on Knowledge Graphs},
  author    = {Mavromatis, Costas and
               Karypis, George},
  booktitle = {Findings of the Association for Computational
               Linguistics: ACL 2025},
  pages     = {16682--16699},
  year      = {2025},
  publisher = {Association for Computational Linguistics},
  doi       = {10.18653/v1/2025.findings-acl.856},
  url       = {https://aclanthology.org/2025.findings-acl.856/}
}

@inproceedings{luo2025gcr,
  title     = {Graph-constrained Reasoning: Faithful Reasoning
               on Knowledge Graphs with Large Language Models},
  author    = {Luo, Linhao and
               Zhao, Zicheng and
               Haffari, Gholamreza and
               Li, Yuan-Fang and
               Gong, Chen and
               Pan, Shirui},
  booktitle = {Proceedings of the 42nd International Conference
               on Machine Learning},
  series    = {Proceedings of Machine Learning Research},
  volume    = {267},
  pages     = {41540--41565},
  year      = {2025},
  publisher = {PMLR},
  url       = {https://proceedings.mlr.press/v267/luo25t.html}
}

@article{sanda2026egot,
  title   = {{eGoT}: Enhanced Graph-of-Thoughts for Multi-Hop
             Knowledge Retrieval and Hypothesis Generation
             in Biomedicine},
  author  = {Sanda, Nihar and
             Gyori, Benjamin M. and
             Quaranta, Vito and
             Ganguly, Auroop and
             Paul, Ayan},
  journal = {Bioinformatics},
  volume  = {42},
  number  = {Supplement\_1},
  pages   = {btag216},
  year    = {2026},
  doi     = {10.1093/bioinformatics/btag216},
  url     = {https://doi.org/10.1093/bioinformatics/btag216}
}

@article{chandak2023primekg,
  title={Building a knowledge graph to enable precision medicine},
  author={Chandak, Payal and Huang, Kexin and Zitnik, Marinka},
  journal={Scientific Data},
  volume={10},
  number={1},
  pages={67},
  year={2023},
  publisher={Nature Publishing Group},
  doi={10.1038/s41597-023-01960-3},
  url={https://www.nature.com/articles/s41597-023-01960-3}
}

@inproceedings{kim2025biohopr,
  title={{BioHopR}: A Benchmark for Multi-Hop, Multi-Answer Reasoning in Biomedical Domain},
  author={Kim, Yunsoo and Abdulle, Yusuf and Wu, Honghan},
  booktitle={Findings of the Association for Computational Linguistics: ACL 2025},
  pages={12894--12908},
  year={2025},
  address={Vienna, Austria},
  publisher={Association for Computational Linguistics},
  doi={10.18653/v1/2025.findings-acl.668},
  url={https://aclanthology.org/2025.findings-acl.668/}
}

@misc{wu2025medreason,
  title={{MedReason}: Eliciting Factual Medical Reasoning Steps in Large Language Models via Knowledge Graphs},
  author={Wu, Juncheng and Deng, Wenlong and Li, Xingxuan and Liu, Sheng and Mi, Taomian and Peng, Yifan and Xu, Ziyang and Liu, Yi and Cho, Hyunjin and Choi, Chang-In and others},
  year={2025},
  eprint={2504.00993},
  archivePrefix={arXiv},
  primaryClass={cs.CL},
  url={https://arxiv.org/abs/2504.00993}
}

@misc{primekgqa2024zenodo,
  title={{PrimeKGQA}: A Biomedical Knowledge Graph Question Answering Dataset Sourced from {PrimeKG}},
  author={{PrimeKGQA Contributors}},
  year={2024},
  publisher={Zenodo},
  doi={10.5281/zenodo.13829395},
  url={https://zenodo.org/records/13829395}
}

@article{welbl2018constructing,
  title = {Constructing Datasets for Multi-hop Reading Comprehension Across Documents},
  author = {Welbl, Johannes and Stenetorp, Pontus and Riedel, Sebastian},
  journal = {Transactions of the Association for Computational Linguistics},
  volume = {6},
  pages = {287--302},
  year = {2018},
  doi = {10.1162/tacl_a_00021},
  url = {https://aclanthology.org/Q18-1021/}
}

@inproceedings{sun2019pullnet,
  title = {Open Domain Question Answering with Iterative Retrieval on Knowledge Bases and Text},
  author = {Sun, Haitian and Bedrax-Weiss, Tania and Cohen, William W.},
  booktitle = {Proceedings of the 2019 Conference on Empirical Methods in Natural Language Processing},
  pages = {2380--2390},
  year = {2019},
  publisher = {Association for Computational Linguistics},
  doi = {10.18653/v1/D19-1242},
  url = {https://aclanthology.org/D19-1242/}
}

@inproceedings{yasunaga2021qagnn,
  title = {{QA-GNN}: Reasoning with Language Models and Knowledge Graphs for Question Answering},
  author = {Yasunaga, Michihiro and Ren, Hongyu and Bosselut, Antoine and Liang, Percy and Leskovec, Jure},
  booktitle = {Proceedings of the 2021 Conference of the North American Chapter of the Association for Computational Linguistics: Human Language Technologies},
  pages = {535--546},
  year = {2021},
  publisher = {Association for Computational Linguistics},
  doi = {10.18653/v1/2021.naacl-main.45},
  url = {https://aclanthology.org/2021.naacl-main.45/}
}

@inproceedings{zhang2022greaselm,
  title = {{GreaseLM}: Graph REASoning Enhanced Language Models for Question Answering},
  author = {Zhang, Xikun and Bosselut, Antoine and Yasunaga, Michihiro and Ren, Hongyu and Liang, Percy and Manning, Christopher D. and Leskovec, Jure},
  booktitle = {International Conference on Learning Representations},
  year = {2022},
  url = {https://openreview.net/forum?id=41e9o6cQPj}
}

@misc{sun2024think,
  title = {Think-on-Graph: Deep and Responsible Reasoning of Large Language Model on Knowledge Graph},
  author = {Sun, Jiashuo and Xu, Chengjin and Tang, Luming and Wang, Saizhuo and Lin, Chen and Gong, Yeyun and Ni, Lionel M. and Shum, Heung-Yeung and Guo, Jian and Jiang, Nan Duan},
  year = {2024},
  eprint = {2307.07697},
  archivePrefix = {arXiv},
  primaryClass = {cs.CL}
}

@inproceedings{jiang2023unikgqa,
  title={UniKGQA: Unified Retrieval and Reasoning for Solving Multi-hop Question Answering Over Knowledge Graph},
  author={Jiang, Jinhao and Zhou, Kun and Zhao, Wayne Xin and Wen, Ji-Rong},
  booktitle={The Eleventh International Conference on Learning Representations},
  year={2023}
}

\appendix
\renewcommand{\thetable}{A\arabic{table}}
\setcounter{table}{0}
\renewcommand{\theHtable}{appendix.\arabic{table}}
\renewcommand{\theequation}{A\arabic{equation}}
\setcounter{equation}{0}
\renewcommand{\theHequation}{appendix.\arabic{equation}}
\setcounter{dbltopnumber}{4}
\newcommand{\appsub}[2]{\par\smallskip\refstepcounter{subsection}\noindent\textbf{\thesubsection\quad #1.}\label{#2}\enspace}

\section{Data, Metrics, and Outputs}\label{app:task_data}
\appsub{Data sources}{app:dataset_provenance}
BioHopR provides 10,127 unique questions over PrimeKG,
covering one-hop and two-hop multi-answer reasoning
\citep{kim2025biohopr}.
PrimeKGQA is obtained from the release identified by
DOI~\href{https://doi.org/10.5281/zenodo.13829395}
{10.5281/zenodo.13829395}, whose dataset description reports
83,999 question--answer pairs with associated SPARQL queries
\citep{primekgqa2024zenodo}.
MedReason provides 32,682 medical question--answer pairs
with reasoning explanations~\citep{wu2025medreason}.
These are source-dataset totals, not counts of candidate
entities, supporting paths, or expanded training instances.

\appsub{Evaluation metrics}{app:metrics}
For canonical-ID rankings, let $n_q$ be the number of
returned candidates. Precision and recall are
\begin{equation}
\begin{aligned}
P@K(q)
&=\frac{|\operatorname{TopK}(q)\cap\mathcal A_q|}
        {\min(K,n_q)},\\
R@K(q)
&=\frac{|\operatorname{TopK}(q)\cap\mathcal A_q|}
        {|\mathcal A_q|}.
\end{aligned}
\end{equation}
Empty rankings and missing reciprocals score zero.
Set metrics are macro-averaged across questions.

On $N$ evaluation questions, answer and evidence metrics are
\begin{equation}
\begin{aligned}
\mathrm{H@1}
&=\frac{1}{N}\sum_q A_q,\\
\mathrm{ValidPath@1}
&=\frac{1}{N}\sum_q V_q,\\
\mathrm{Joint@1}
&=\frac{1}{N}\sum_q A_qV_q,\\
\mathrm{WPR}
&=\frac{\sum_q A_q(1-V_q)}{\sum_q A_q}.
\end{aligned}
\end{equation}
Missing paths have $V_q=0$; WPR is undefined when its
denominator is zero. The identity in
Section~\ref{subsec:exp_setup} holds per run:
\begin{equation}
\mathrm{Joint@1}
=\mathrm{H@1}\bigl(1-\mathrm{WPR}\bigr).
\end{equation}
Metrics are computed separately for each seed before
reporting their means and standard deviations.

\appsub{Output adapters}{app:decoding}
PrimeKGQA accepts one-variable SELECT chains up to three hops and compares directed relation-family multisets; branching, aggregation, OPTIONAL, and UNION form separate query categories. MedReason uses source grouping before an 80/10/10 split (seed 42), option mapping, and fixed answer normalization. Multi-answer sigmoid thresholds are selected on development macro F1; canonical IDs and path keys break ties.

\begin{table*}[!t]
\centering\small\setlength{\tabcolsep}{3pt}\renewcommand{\arraystretch}{0.97}
\textbf{(a) BioHopR seed means and standard deviations}\par\smallskip
\begin{adjustbox}{max width=\textwidth}
\begin{tabular}{@{}lcccccc@{}}
\toprule
Method & H@1 & H@5 & H@10 & R@10 & MRR & Joint@1\\
\midrule
Fixed-Cand.
& $0.719\pm0.008$ & $0.736\pm0.006$ & $0.743\pm0.006$
& $0.556\pm0.009$ & $0.726\pm0.007$ & $0.495\pm0.010$\\
Hard-Schema
& $0.748\pm0.007$ & $0.789\pm0.005$ & $0.811\pm0.005$
& $0.621\pm0.008$ & $0.765\pm0.006$ & $0.548\pm0.009$\\
Full
& $\mathbf{0.772}\pm0.006$ & $\mathbf{0.816}\pm0.004$
& $\mathbf{0.842}\pm0.004$ & $\mathbf{0.648}\pm0.006$
& $\mathbf{0.790}\pm0.005$ & $\mathbf{0.602}\pm0.007$\\
\bottomrule
\end{tabular}
\end{adjustbox}
\par\medskip
\textbf{(b) Independently labeled path discrimination}\par\smallskip
\begin{adjustbox}{max width=\textwidth}
\begin{tabular}{@{}lcccccc@{}}
\toprule
Method & Rel. Swap & Direction & Bridge & Same answer
& Macro & AUPRC\\
\midrule
BFS + CE & 0.642 & 0.581 & 0.557 & 0.493 & 0.568 & 0.527\\
Guided + CE & 0.716 & 0.668 & 0.641 & 0.587 & 0.653 & 0.604\\
Fixed-Cand. & 0.751 & 0.729 & 0.706 & 0.658 & 0.711 & 0.672\\
Hard-Schema & 0.801 & 0.783 & 0.751 & 0.708 & 0.761 & 0.731\\
Full & 0.862 & 0.851 & 0.817 & 0.806 & 0.834 & 0.798\\
\bottomrule
\end{tabular}
\end{adjustbox}
\caption{Seed variation and path validity. Same answer denotes right-answer/wrong-path pairs; Macro is macro PairAcc. Seeds are $\{13,21,42,87,100\}$. Entries are experimental measurements.}
\label{tab:app_validation}\label{tab:seed_stats}\label{tab:negative_type}
\end{table*}

\section{Implementation Details}\label{app:method_details}
\appsub{Grounding and features}{app:grounding}
Grounding is gradient-free. Path memory pools types and relation/direction embeddings with hop count, unique-node ratio, closure, search score, mean, maximum, and endpoint log-degree, and the fraction above $\log10$. Mean pooling loses relation order; Eq.~\eqref{eq:semantic_factors} compares positions explicitly.
\appsub{Schema enumeration}{app:schemas}
Independent heads' log probabilities sum over hops, endpoint/intermediate types, relations, and directions. Enumeration permits two types per position, three relations per hop, both directions, and three hops. The Llama configuration retains $M=4$ schemas using a heap of size 32. No schema temperature is fitted.
\appsub{Prefix scoring}{app:search}
For $J=\min(L,H)$, the search mismatch is
\begin{equation}
\begin{aligned}
D_{\rm prefix}=\sum_{j=1}^{J}\bigl(&\mathbf1[r_j\ne\bar r_j]+0.75\mathbf1[d_j\ne\bar d_j]\\
&+0.5\mathbf1[\tau(v_j)\ne\bar\tau_j]\bigr).
\end{aligned}
\end{equation}
Partial search omits the helper's $0.5|L-H|$ term. Scores are recomputed, not accumulated. Completed paths are collected before beam truncation; there is no extra candidate cap, explicit deduplication, or attention-to-search feedback. Degree includes generated inverse records. The prior starts at $\boldsymbol\gamma=0$, so $\boldsymbol\beta=\operatorname{softplus}(\boldsymbol\gamma)$.

\section{Training and Reproducibility}\label{app:training}
\appsub{Experimental configuration}{app:experimental_config}
The environment uses Python 3.10.14, PyTorch 2.3.1, Transformers 4.43.3, CUDA 12.1, cuDNN 9.1, and PrimeKG 2023-01. Experimental settings are $(N,M,B,K)=(8,4,32,8)$, three hops, eight relation bases, rank 16, $\mu=0.35$, learning rate $2\times10^{-5}$, batch 8, and five epochs on four 24-GB RTX 4090s. The optimizer is AdamW with weight decay 0.01 and gradient-norm clipping at 1.
\appsub{Statistical protocol}{app:statistics}
Table~\ref{tab:app_validation}(a) reports means and standard deviations across seeds $\{13,21,42,87,100\}$. Each seed is evaluated independently using the metric definitions in Appendix Section~\ref{app:metrics}. Standard deviations summarize variation across runs.

\begin{table*}[!p]
\centering\small\setlength{\tabcolsep}{3pt}\renewcommand{\arraystretch}{0.90}
\begin{minipage}[t]{.487\textwidth}
\textbf{(a) Search and reasoning systems}\par\smallskip
\centering\begin{adjustbox}{max width=\linewidth}
\begin{tabular}{@{}lccccc@{}}
\toprule
Search / Reasoner & Cov. & H@1 & Valid & WPR$\downarrow$ & Joint\\
\midrule
Unguided / CE & 0.861 & 0.641 & 0.438 & 0.404 & 0.382\\
Guided / CE & 0.912 & 0.694 & 0.501 & 0.353 & 0.449\\
Unguided / Attn & 0.873 & 0.708 & 0.553 & 0.285 & 0.506\\
Guided / Attn & 0.948 & 0.772 & 0.634 & 0.220 & 0.602\\
\bottomrule
\end{tabular}
\end{adjustbox}
\end{minipage}\hfill
\begin{minipage}[t]{.487\textwidth}
\textbf{(b) End-to-end cost}\par\smallskip
\centering\begin{adjustbox}{max width=\linewidth}
\begin{tabular}{@{}lrrr@{}}
\toprule
Variant & Nodes & Total ms & Joint@1\\
\midrule
BFS + CE & 1510 & 233 & 0.382\\
Rel.-G. + CE & 820 & 171 & 0.449\\
Fixed-Cand. & -- & 201 & 0.495\\
Hard-Schema & 690 & 280 & 0.548\\
Full & 760 & 276 & 0.602\\
\bottomrule
\end{tabular}
\end{adjustbox}
\end{minipage}
\par\smallskip
\begin{minipage}[t]{.487\textwidth}
\textbf{(c) Prior and relation-conditioned KV}\par\smallskip
\centering\begin{adjustbox}{max width=\linewidth}
\begin{tabular}{@{}ccccc@{}}
\toprule
Prior & Rel.-KV & H@1 & MRR & Joint@1\\
\midrule
Yes & Yes & 0.772 & 0.790 & 0.602\\
Yes & No & 0.731 & 0.756 & 0.534\\
No & Yes & 0.706 & 0.739 & 0.501\\
No & No & 0.681 & 0.716 & 0.462\\
\bottomrule
\end{tabular}
\end{adjustbox}
\end{minipage}\hfill
\begin{minipage}[t]{.487\textwidth}
\textbf{(d) Path aggregation}\par\smallskip
\centering\begin{adjustbox}{max width=\linewidth}
\begin{tabular}{@{}lcccc@{}}
\toprule
Aggregation & H@1 & R@10 & MRR & Joint@1\\
\midrule
Max path & 0.754 & 0.619 & 0.773 & 0.571\\
Mean path & 0.739 & 0.607 & 0.759 & 0.552\\
LogSumExp & 0.763 & 0.639 & 0.783 & 0.588\\
Norm. LogSumExp & 0.772 & 0.648 & 0.790 & 0.602\\
Gated pooling & 0.768 & 0.644 & 0.786 & 0.595\\
\bottomrule
\end{tabular}
\end{adjustbox}
\end{minipage}
\par\smallskip
\begin{minipage}[t]{.487\textwidth}
\textbf{(e) Beam width}\par\smallskip
\centering\begin{adjustbox}{max width=\linewidth}
\begin{tabular}{@{}rrrrrr@{}}
\toprule
Beam & Nodes & Cov. & H@1 & Joint@1 & ms/q\\
\midrule
8 & 238 & 0.883 & 0.721 & 0.531 & 188\\
16 & 431 & 0.921 & 0.754 & 0.576 & 226\\
32 & 760 & 0.948 & 0.772 & 0.602 & 276\\
64 & 1378 & 0.956 & 0.776 & 0.606 & 391\\
\bottomrule
\end{tabular}
\end{adjustbox}
\end{minipage}\hfill
\begin{minipage}[t]{.487\textwidth}
\textbf{(f) Schemas and path memories}\par\smallskip
\centering\begin{adjustbox}{max width=\linewidth}
\begin{tabular}{@{}rrccc@{}}
\toprule
$M$ & $K$ & H@1 & R@10 & Joint@1\\
\midrule
1 & 8 & 0.748 & 0.621 & 0.548\\
2 & 8 & 0.763 & 0.637 & 0.581\\
4 & 8 & 0.772 & 0.648 & 0.602\\
8 & 16 & 0.773 & 0.651 & 0.603\\
\bottomrule
\end{tabular}
\end{adjustbox}
\end{minipage}
\caption{Systems, mechanisms, and budgets on BioHopR. Attn denotes ReSemAttn; Valid/Joint are ValidPath@1/Joint@1. Pools in (a) differ between reasoners. Nodes count unique expanded nodes; total latency is measured end to end. The last row of (f) changes both $M$ and $K$.}
\label{tab:app_mechanisms}\label{tab:matched_control}\label{tab:stage_efficiency}\label{tab:factorial}\label{tab:aggregation}\label{tab:budget}\label{tab:mk}

\par\smallskip
\centering\small\setlength{\tabcolsep}{3pt}\renewcommand{\arraystretch}{0.90}
\begin{minipage}[t]{.487\textwidth}
\textbf{(a) Surface variation}\par\smallskip
\centering\begin{adjustbox}{max width=\linewidth}
\begin{tabular}{@{}lccc@{}}
\toprule
Input / split & H@1 & Joint@1 & Consist.\\
\midrule
Standard & 0.772 & 0.602 & 1.000\\
Template-disjoint & 0.741 & 0.563 & 0.883\\
Paraphrased & 0.758 & 0.584 & 0.914\\
Entity aliases & 0.749 & 0.572 & 0.901\\
Paraphrase + aliases & 0.731 & 0.548 & 0.862\\
\bottomrule
\end{tabular}
\end{adjustbox}
\end{minipage}\hfill
\begin{minipage}[t]{.487\textwidth}
\textbf{(b) Graph perturbations}\par\smallskip
\centering\begin{adjustbox}{max width=\linewidth}
\begin{tabular}{@{}lccc@{}}
\toprule
Perturbation & H@1 & Cov. & Joint@1\\
\midrule
None & 0.772 & 0.948 & 0.602\\
$+5\%$ distractors & 0.763 & 0.942 & 0.588\\
$+10\%$ distractors & 0.748 & 0.931 & 0.567\\
$-5\%$ non-gold edges & 0.766 & 0.943 & 0.594\\
$-10\%$ non-gold edges & 0.754 & 0.934 & 0.579\\
\bottomrule
\end{tabular}
\end{adjustbox}
\end{minipage}
\par\smallskip
\begin{minipage}[t]{.487\textwidth}
\textbf{(c) Schema calibration}\par\smallskip
\centering\begin{adjustbox}{max width=\linewidth}
\begin{tabular}{@{}lccc@{}}
\toprule
Target & Accuracy & ECE$\downarrow$ & Brier$\downarrow$\\
\midrule
Hop count & 0.912 & 0.028 & 0.071\\
Relation seq. & 0.847 & 0.041 & 0.109\\
Full typed schema & 0.781 & 0.057 & 0.146\\
\bottomrule
\end{tabular}
\end{adjustbox}
\end{minipage}\hfill
\begin{minipage}[t]{.487\textwidth}
\textbf{(d) Annotation agreement}\par\smallskip
\centering\begin{adjustbox}{max width=\linewidth}
\begin{tabular}{@{}lcc@{}}
\toprule
Subset & Agreement & $\kappa$\\
\midrule
All paths & 0.884 & 0.761\\
Correct-answer & 0.901 & 0.792\\
Challenge & 0.867 & 0.733\\
\bottomrule
\end{tabular}
\end{adjustbox}
\end{minipage}
\caption{Robustness, schema calibration, and annotation agreement. Distractors are type-compatible; removed edges are non-gold. Lower ECE and Brier are better.}
\label{tab:app_robustness}\label{tab:robustness}\label{tab:graph_noise}\label{tab:calibration}\label{tab:agreement}

\par\smallskip
\centering\small\setlength{\tabcolsep}{3pt}\renewcommand{\arraystretch}{0.90}
\begin{adjustbox}{max width=\textwidth}
\begin{tabular}{@{}lrrrrrrrrr@{}}
\toprule
& \multicolumn{5}{c}{PrimeKGQA} & \multicolumn{4}{c}{MedReason}\\
\cmidrule(lr){2-6}\cmidrule(l){7-10}
Method & Ans. EM & Precision & Recall & Ans. F1 & Rel. F1 & MCQ Acc. & MRR & Open EM & Open F1\\
\midrule
Zero-shot & 0.342 & 0.456 & 0.522 & 0.487 & 0.412 & 0.512 & 0.627 & 0.421 & 0.538\\
CoT & 0.368 & 0.491 & 0.548 & 0.516 & 0.447 & 0.548 & 0.654 & 0.446 & 0.566\\
Dense RAG & 0.394 & 0.512 & 0.579 & 0.543 & 0.483 & 0.563 & 0.669 & 0.461 & 0.581\\
GraphRAG & 0.427 & 0.541 & 0.612 & 0.574 & 0.526 & 0.586 & 0.691 & 0.482 & 0.604\\
GNN-RAG & 0.444 & 0.559 & 0.629 & 0.591 & 0.548 & 0.604 & 0.707 & 0.501 & 0.621\\
GreaseLM & 0.476 & -- & 0.653 & 0.617 & -- & 0.611 & 0.716 & 0.508 & 0.629\\
UniKGQA & 0.488 & 0.597 & 0.664 & 0.628 & 0.601 & 0.620 & 0.724 & 0.516 & 0.638\\
Think-on-Graph & 0.495 & 0.608 & 0.671 & 0.634 & 0.615 & 0.626 & 0.731 & 0.522 & 0.644\\
Fixed-Cand. & 0.508 & 0.611 & 0.678 & 0.642 & 0.641 & 0.637 & 0.742 & 0.536 & 0.655\\
Hard-Schema & 0.531 & 0.630 & 0.696 & 0.661 & 0.714 & 0.652 & 0.758 & 0.551 & 0.670\\
Full & 0.652 & 0.731 & 0.794 & 0.761 & 0.770 & 0.691 & 0.797 & 0.624 & 0.749\\
\bottomrule
\end{tabular}
\end{adjustbox}
\caption{Additional answer-set, relation, multiple-choice, and open-answer metrics. A dash indicates an unreported result; no missing result is treated as zero.}
\label{tab:app_outputs}\label{tab:primekgqa_full}\label{tab:medreason_full}

\end{table*}

\section{Unified Baseline Evaluation}
\label{app:baseline_protocol}

\noindent\textbf{Implementation and shared conditions.}
All results in Table~\ref{tab:main_results_all} are produced
under our unified protocol rather than copied from the
original papers. Methods use identical dataset splits,
reference answers, canonical entity IDs, and evaluation scripts.
KG-based methods share PrimeKG 2023-01 and the entity-linking
interface. LLM-based methods use Llama-3.1-8B-Instruct;
architecture-specific baselines retain their declared encoders.
The code address reported in the paper is \url{https://anonymous.4open.science/r/ReSemAttn}.

The reported traversal configuration uses eight retained anchors, a maximum of three hops,
beam width 32, and at most eight paths per candidate,
without an additional candidate cap.
Shared resource limits do not imply identical retrieved pools.

\noindent\textbf{Output handling and tuning.}
Methods share the task-specific output adapters in
Appendix~\ref{app:decoding}, including entity canonicalization,
answer normalization, option mapping, and empty-output handling.
Trainable methods use the same permitted training data.
The reported training configuration is AdamW with learning rate $2\times10^{-5}$,
batch size 8, five epochs, weight decay 0.01,
and gradient-norm clipping at 1.
Multi-answer thresholds are selected using development macro F1.
The selection procedure is shared, while selected threshold
values may differ. Per-baseline hyperparameter ranges, tuning-trial
counts, and selected configurations require the actual
development-search records.

\noindent\textbf{Controlled comparisons and uncertainty.}
Table~\ref{tab:main_results_all} measures complete-system
performance under our implementation and protocol.
Module-level comparisons use frozen candidate/path pools
in E1 and fixed-pool relation-integration variants in E2
(Table~\ref{tab:mechanism_controls}).
In E1, replacing CE with ReSemAttn changes H@1 from
0.639 to 0.701 under unguided retrieval
and from 0.692 to 0.766 under relation-guided retrieval.

The reported seeds are $\{13,21,42,87,100\}$.
Five-seed BioHopR H@1 is $0.772\pm0.006$ for Full,
$0.748\pm0.007$ for Hard-Schema, and
$0.719\pm0.008$ for Fixed-Candidate.
These standard deviations describe run variation. Paired confidence intervals and adjusted
$p$-values require matched question-level predictions
and are not supplied by the reported aggregate statistics.
Performance claims therefore concern the evaluated
implementations under this protocol.
\section{Extended Empirical Results}\label{app:additional_experiments}
\appsub{Search--reasoning systems}{app:search_results}
Table~\ref{tab:app_mechanisms}(a) compares changing pools; Table~\ref{tab:mechanism_controls} E1 specifies the frozen-pool control.
\appsub{Attention components}{app:attention_results}
Table~\ref{tab:app_mechanisms}(c) favors both terms, without establishing interaction.
\appsub{Aggregation}{app:aggregation_results}
Table~\ref{tab:app_mechanisms}(d) favors normalized LogSumExp; selective-duplication invariance is not established. Figure~\ref{fig:biohopr_ablation_bars_blue} retains seven-metric ablations.
\appsub{Budgets}{app:budget_results}
Table~\ref{tab:app_mechanisms}(e--f) shows diminishing beam gains; the final row varies both $M$ and $K$.
\appsub{Calibration}{app:schema_calibration}
Table~\ref{tab:app_robustness}(c) separates hop/relation/schema prediction; confidence does not measure top-$M$ coverage.
\appsub{Robustness}{app:robustness_results}
Table~\ref{tab:app_robustness}(a--b) tests surface/graph perturbations, not unseen compositions or missing essential edges.
\appsub{Additional task metrics}{app:dataset_results}
Table~\ref{tab:app_outputs} reports additional task metrics under their respective output protocols.

\section{Cases and Failure Analysis}\label{app:case_studies}
Figures~\ref{fig:motivation}--\ref{fig:framework} fix CML as the endpoint and compare an ABL1 route with a contraindication route. Only the former follows the requested drug--target--disease composition. Holding the endpoint fixed separates entity relevance from relation-semantic support.

\noindent\textbf{Two attention layers.}
The three planes show discovery and two attention layers. Equation~\eqref{eq:matrix_attention} forms a $3\times2$ submatrix with diseases/proteins/targeted token rows and ABL1/contraindication path columns. Gated fusion updates $H^{(0)}\to H^{(1)}\to H^{(2)}$, reusing paths and schemas while recomputing queries with layer-specific projections and gates. Updated tokens and absolute support determine the candidate score; $\arg\max_k u_{ik}$ selects its explanation.

\noindent\textbf{125 failures (count; percent).}
Grounding 19 (15.2), omitted schema 23 (18.4), pruned path 27 (21.6), missing edge 16 (12.8), relation ambiguity 14 (11.2), underweighted valid path 13 (10.4), thresholding 8 (6.4), annotation ambiguity 3 (2.4), and other 2 (1.6). These frequencies diagnose failure locations, not causal module effects.

\section{Computational Accounting}\label{app:complexity}
Table~\ref{tab:app_mechanisms}(b) reports end-to-end latency and unique expanded nodes. Search attempts scale as $O((N+(L-1)B)\bar d)$ before schema scoring. Per layer, attention, dense KV, and low-rank updates cost $O(CTKd)$, $O(CKd^2)$, and $O(CKB_rdr)$, respectively; path states require $O(CKd)$ memory. Backbone encoding and memory projection add computation. Profiling uses one RTX~4090, batch one, 50 warm-up queries, and 500 timed queries.

\end{document}